\RequirePackage{fix-cm}
\documentclass[twocolumn]{svjour3}          
\smartqed  
\usepackage{graphicx}%
\usepackage{multirow}%
\usepackage{amsmath,amssymb,amsfonts}%
\usepackage{array}
\usepackage{mathrsfs}%
\usepackage[title]{appendix}%
\usepackage{xcolor}%
\usepackage{textcomp}%
\usepackage{manyfoot}%
\usepackage{booktabs}%
\usepackage{algorithmicx}%
\usepackage{algpseudocode}%
\usepackage{listings}%

\usepackage{array}
\usepackage{stfloats}
\usepackage{url}
\usepackage{verbatim}
\usepackage{graphics} 
\usepackage{algpseudocode}
\usepackage{listings}
\usepackage{xcolor}
\usepackage{float}
\usepackage{lettrine}
\usepackage{subcaption}
\usepackage{arydshln}
\usepackage{natbib}
\usepackage[bookmarks=true,colorlinks=true,urlcolor=blue,citecolor=blue]{hyperref}

\usepackage[ruled,vlined]{algorithm2e}
\usepackage[10pt]{moresize}
\usepackage{threeparttable}

\definecolor{codegreen}{rgb}{0,0.6,0}
\definecolor{codegray}{rgb}{0.5,0.5,0.5}
\definecolor{codepurple}{rgb}{0.58,0,0.82}
\definecolor{backcolour}{rgb}{0.95,0.95,0.92}

\definecolor{cr}{rgb}{1,0,0}
\definecolor{cg}{rgb}{0,0.62,0}
\definecolor{cb}{rgb}{0,0,1}

\lstdefinestyle{mystyle}{
    backgroundcolor=\color{backcolour},   
    commentstyle=\color{codegreen},
    keywordstyle=\color{magenta},
    numberstyle=\tiny\color{codegray},
    stringstyle=\color{codepurple},
    basicstyle=\ttfamily\scriptsize,
    breakatwhitespace=false,         
    breaklines=true,                 
    captionpos=b,                    
    keepspaces=true,                 
    numbers=left,                    
    numbersep=5pt,                  
    showspaces=false,                
    showstringspaces=false,
    inputencoding=utf8,
    extendedchars=false,
    showtabs=false,                  
    tabsize=2
}
\newcommand{\PAR}[1]{\vskip4pt \noindent{\bf #1~}}

\usepackage[normalem]{ulem}
\useunder{\uline}{\ul}{}

\makeatletter
\renewcommand\paragraph{
  \@startsection{paragraph} 
  {4} 
  {\z@} 
  {.5em \@plus1ex \@minus.2ex} 
  {-.5em} 
  {\normalfont\normalsize\bfseries} 
}
\makeatother
\begin{document}
\sloppy

\title{EdgePoint2: Compact Descriptors for Superior Efficiency and Accuracy 
}


\author{Haodi Yao \and
        Fenghua He \and
        Ning Hao \and
        Chen Xie
}


\institute{Haodi Yao \at
                School of Astronautics, Harbin Institute of Technology, China \\
                \email{20B904013@stu.hit.edu.cn}
           \and
           Fenghua He \at
                School of Astronautics, Harbin Institute of Technology, China \\
                \email{hefenghua@hit.edu.cn}
           \and
           Ning Hao \at
                School of Astronautics, Harbin Institute of Technology, China \\
                \email{haoning0082022@163.com}
           \and
           Xie Chen \at
                School of Astronautics, Harbin Institute of Technology, China \\
                \email{2048872092@qq.com}
}

\date{}

\maketitle

\begin{abstract}
The field of keypoint extraction, which is essential for vision applications like Structure from Motion (SfM) and Simultaneous Localization and Mapping (SLAM), has evolved from relying on handcrafted methods to leveraging deep learning techniques. While deep learning approaches have significantly improved performance, they often incur substantial computational costs, limiting their deployment in real-time edge applications. Efforts to create lightweight neural networks have seen some success, yet they often result in trade-offs between efficiency and accuracy. Additionally, the high-dimensional descriptors generated by these networks poses challenges for distributed applications requiring efficient communication and coordination, highlighting the need for compact yet competitively accurate descriptors. In this paper, we present EdgePoint2, a series of lightweight keypoint detection and description neural networks specifically tailored for edge computing applications on embedded system. The network architecture is optimized for efficiency without sacrificing accuracy. To train compact descriptors, we introduce a combination of Orthogonal Procrustes loss and similarity loss, which can serve as a general approach for hypersphere embedding distillation tasks. Additionally, we offer 14 sub-models to satisfy diverse application requirements. Our experiments demonstrate that EdgePoint2 consistently achieves state-of-the-art (SOTA) accuracy and efficiency across various challenging scenarios while employing lower-dimensional descriptors (32/48/64). Beyond its accuracy, EdgePoint2 offers significant advantages in flexibility, robustness, and versatility. Consequently, EdgePoint2 emerges as a highly competitive option for visual tasks, especially in contexts demanding adaptability to diverse computational and communication constraints.\footnote{Demo and weights are available at \url{https://github.com/HITCSC/EdgePoint2}.}
\end{abstract}


\section{Introduction}

Keypoint extraction, which involves both keypoint detection and description, plays an essential role in vision applications such as Structure from Motion (SfM)~\citep{song2015high, glomap} and Simultaneous Localization and Mapping (SLAM)~\citep{ORBSLAM3, MAVIS}. Given the critical importance of these tasks in robotics and UAVs, they are frequently implemented on embedded systems. Previous works depended on handcrafted methods~\citep{SIFT, ORB, FAST, BRIEF}. While these methods have been used wildly in applications~\citep{ORBSLAM3, colmap1, VINS-Mono, KimeraMulti2022}, they are still encountering significant challenges regarding accuracy and robustness. Driven by the substantial advancements in deep learning, keypoint extraction has demonstrated remarkable improvements~\citep{LF-Net, SuperPoint, SOSNet}. Several studies, including~\citep{SP-VSLAM, SPSG-Odometry}, have shown improved performance by leveraging these advanced learning-based keypoint extraction methods. But along with the performance, the deep-learning based methods also come with a significant computational burden~\citep{SuperPoint, DISK}, which limits their applicability for real-time applications on embedded systems. This issue becomes increasingly pronounced as embedded devices are required to process data from multiple cameras simultaneously during deployment~\citep{OmniNXT,M2DGR}. 

In order to enable realtime network inference on edge devices, numerous works have proposed various lightweight neural network models specifically designed for enhancing backbone efficiency~\citep{MobileNetv2, MobileNetv3, FasterNet, EfficientNetv2}. Building on these achievements, several studies have attempted to improve the efficiency of keypoint extraction on edge devices~\citep{HF-Net, XFeat}. However, despite these efforts, the resulting networks still tend to be either too resource-intensive for embedded systems or require compromises in accuracy. 

\begin{figure*}[htbp]
    \centering
    \begin{subfigure}[b]{0.48\textwidth}
        \includegraphics[width=\textwidth]{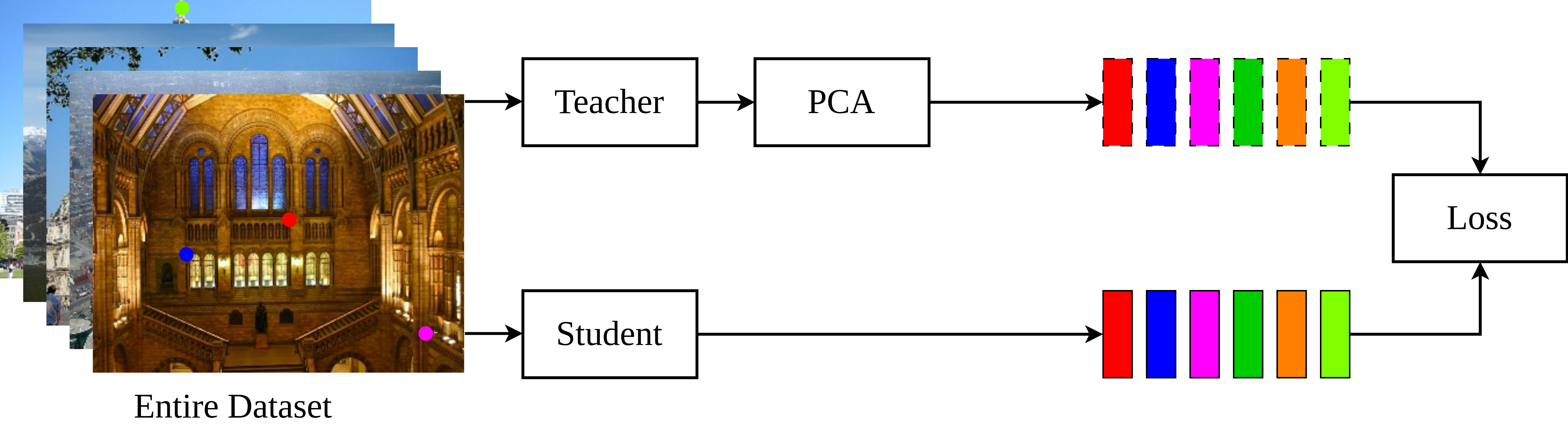}
        \caption{HF-Net~\citep{HF-Net}}
        \label{fig:Fig1-a}
    \end{subfigure}
    \hfill
    \begin{subfigure}[b]{0.48\textwidth}
        \includegraphics[width=\textwidth]{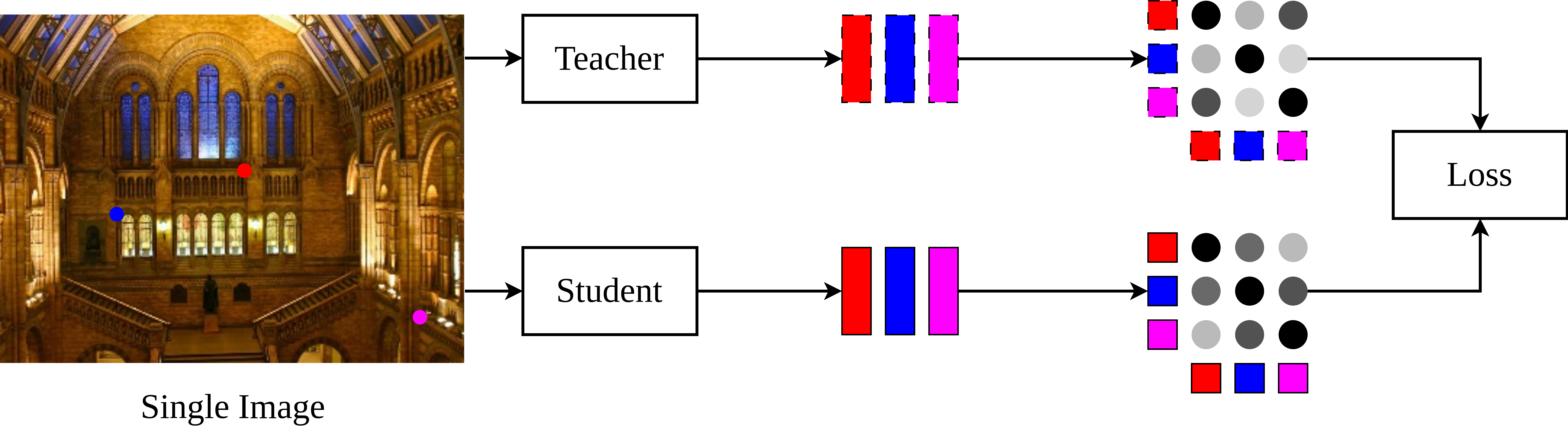}
        \caption{AWDesc~\citep{AWDesc}}
        \label{fig:Fig1-b}
    \end{subfigure}
    \\
    \begin{subfigure}[b]{0.48\textwidth}
        \includegraphics[width=\textwidth]{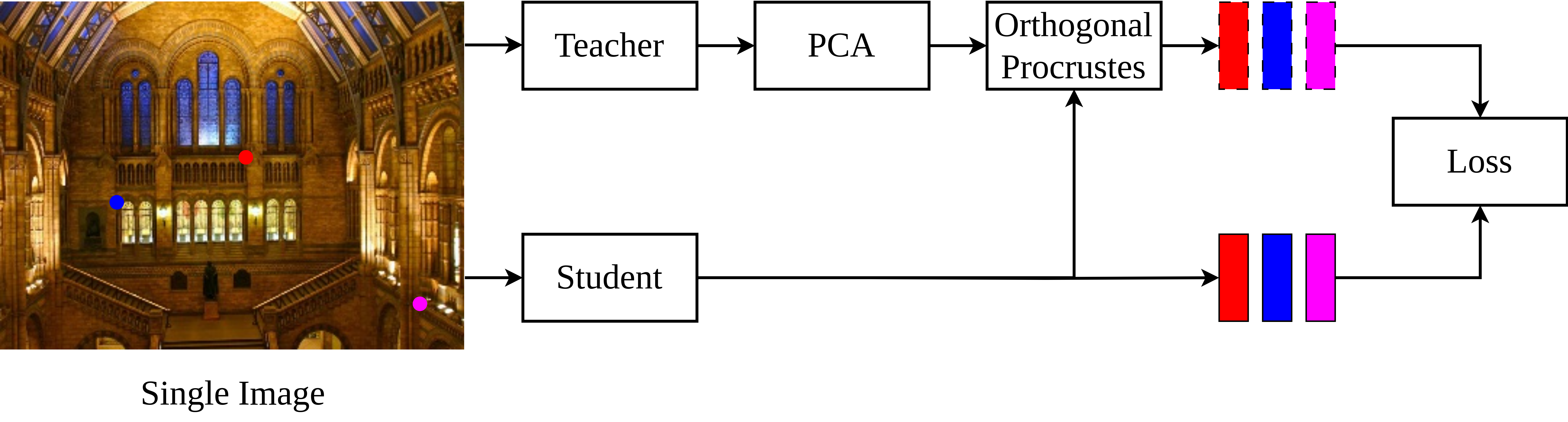}
        \caption{EdgePoint~\citep{EdgePoint}}
        \label{fig:Fig1-c}
    \end{subfigure}
    \hfill
    \begin{subfigure}[b]{0.48\textwidth}
        \includegraphics[width=\textwidth]{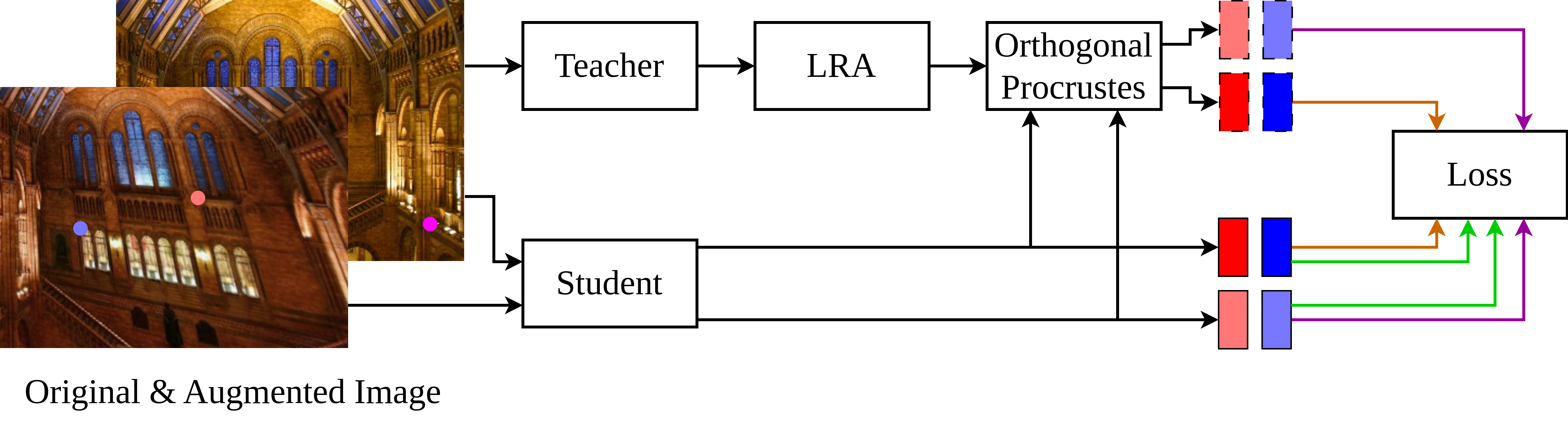}
        \caption{Ours}
        \label{fig:Fig1-d}
    \end{subfigure}
    \caption{\raggedright \textbf{Comparison of local descriptor distillation approaches.} The dashed and solid boxes denote descriptors generated by the teacher and student models, respectively. Specifically, (a) illustrates vanilla direct distillation between feature descriptors; (b) optimizes distribution matching via distance matrix loss computation; (c) shows the Orthogonal Procrustes-based alignment with PCA; (d) presents the proposed method that combines Orthogonal Procrustes loss (incorporating LRA) and similarity loss through image augmentation, resulting in SOTA accuracy.}
    \label{fig:Fig1}
\end{figure*}

Furthermore, for distributed applications such as multi-robot cooperation and collaboration~\citep{D2SLAM, OmniSwarm}, these networks often generate high-dimensional descriptors—typically 256 or 128 dimensions—which places a heavy communication burden and poses challenges for tasks like collaborative localization. While attempts have been made to create lower-dimensional descriptors~\citep{XFeat, SiLK, EdgePoint}, these solutions often struggle to match the accuracy of SOTA methods. Consequently, there is a persistent demand for compact descriptors that not only maintain competitive accuracy but also facilitate efficient communication in distributed applications where bandwidth is limited. Developing such descriptors could significantly enhance the performance and scalability of distributed application systems.

In this work, we propose EdgePoint2, a super lightweight model specifically designed for resource-constrained embedded devices. The model architecture comprises a minimum of 28k parameters and is optimized for real-time applications, whether utilizing CPU or GPU. By separating the detection and description feature maps and selecting the optimal size for each, we improve inference speed while maintaining high performance. For descriptor distillation, current methods struggle to effectively preserving the relative distribution structure of descriptors in a lower-dimensional space. To overcome this challenge, we introduces an innovative descriptor loss function that incorporates Orthogonal Procrustes loss for distillation and similarity loss for self-supervision. We illustrate the distinctions between previous descriptor distillation methods and our approach in Fig. \ref{fig:Fig1}. This advancement enables EdgePoint2 to achieve performance that surpasses the teacher model while requiring fewer computational resources. Furthermore, we offer various model configurations (ranging from tiny to enormous) to accommodate diverse usage requirements. EdgePoint2 achieves SOTA accuracy while employing lower-dimensional descriptors (32/48/64), which is particularly significant for tasks with limited communication bandwidth. To validate the effectiveness of EdgePoint2, we conducted comprehensive evaluations across multiple tasks with different model configurations. These tasks included, but are not limited to, homography estimation (HPatches~\citep{HPatches}), relative pose estimation (MegaDepth-1500~\citep{MegaDepth}, ScanNet-1500~\citep{ScanNet}, IMC2022~\citep{IMC2022}), and visual localization (Aachen Day-Night v1.1~\citep{HLocDataset}, InLoc~\citep{InLoc}). The experimental results demonstrated consistent performance and adaptability in diverse environmental conditions. Remarkably, even our EdgePoint2 32-dim models outperform several SOTA methods that utilize descriptor dimensions of 64 or more. To the best of our knowledge, the EdgePoint2 series offers an optimal performance in efficiency and efficiency. The main contributions are summarized as follows:

\begin{itemize}
    \item We present a novel network architecture that utilizes fewer parameters while reducing computational costs, achieving high inference speed without compromising performance. This architecture employs simple convolutional layers, making it suitable for various acceleration devices and libraries. Notably, our approach can achieve real-time performance even when operating solely on ARM CPUs.
    \item We incorporate the Orthogonal Procrustes loss with the similarity loss to learn compact descriptors through distillation, thereby fully leveraging the pre-allocation of the descriptor space.
    \item We offer a series of models with various configurations and descriptor dimensionalities, providing significant flexibility in selecting the optimal model for various challenges that require robustness, scalability, and adaptability to different computational constraints.
    \item We conduct evaluations across multiple datasets and diverse scenarios. The results demonstrate that EdgePoint2 excels in challenging environments while maintaining consistent performance.
\end{itemize}

\section{Related Work}

\subsection{Lighter and Better Model}

Early deep learning-based keypoint detection and description networks primarily employed VGG-style~\citep{VGG} backbones, as~\citep{SuperPoint, UnsuperPoint, KP2D, D2-Net, DeDoDe, DeDoDev2}. Although these architectures demonstrated the potential of deep learning for feature extraction, they often come with substantial computational demands. In response, recent approaches have increasingly focused on designing lightweight networks.

To balance accuracy and speed, the Feature Pyramid Network was adopted, which leverages multi-scale feature maps to effectively represent objects of varying sizes~\citep{ALIKE, ASLFeat, ALIKED, XFeat}. Some efforts aim to maximize accuracy by generating detection and descriptor maps using features from the original image size~\citep{ALIKE, ALIKED}. These methods utilizes fewer parameters and floating-point operations per second (FLOPs), achieving enhanced accuracy. However, they still require high-performance accelerators such as GPUs. Conversely,~\citep{SiLK, EdgePoint} pursued extremely fast architectures that employ simple convolution layers to attain high speed and low parameter counts. Nonetheless, the performance of these methods is frequently compromised by image size due to their limited receptive fields. Deformable Convolution Networks (DCN) have also been utilized for geometric feature extraction in~\citep{ASLFeat,ALIKED,SeLF}. However, the increased computational complexity and memory overhead associated with DCN limit their effectiveness for inference acceleration on embedded devices.

\subsection{Compact Descriptor}

Early descriptor networks generally generate 256-dimensional descriptors~\citep{SuperPoint,KP2D,UnsuperPoint}. This not only increases computational, storage, and communication burdens but also slows down the matching process. Recent advancements, such as~\citep{R2D2,ASLFeat,ALIKE,ALIKED,AWDesc}, have successfully reduced this dimension to 128 while achieving better performance. Furthermore, studies like~\citep{ALIKE,ALIKED,AWDesc} have proposed 64-dimensional compact descriptors that demonstrate superior performance compared to higher-dimensional descriptors. Some research even advocates for 32-dimensional descriptors, suggesting that low-dimensional descriptors can still perform effectively, although they do not achieve SOTA performance when compared to 64- or 128-dimensional descriptors~\citep{SiLK,EdgePoint}. Additionally, some techniques apply post-processing methods like Principal Component Analysis (PCA) for descriptor compression~\citep{D2SLAM,NetVLAD}. However, the requirement for pre-collected datasets limit their flexibility.

Besides low descriptor dimensions, another type of compact descriptor is boolean descriptors. Previous research on boolean descriptors has primarily focused on global descriptors for visual place recognition and image retrieval~\citep{DeepBit, DCBD-MQ, D-GraphBit}. ZippyPoint~\citep{ZippyPoint} focuses on boolean local descriptors, achieving smaller descriptor sizes and faster matching procedures, but this comes at the cost of accuracy and stability. Furthermore, it requires special optimizations to achieve this acceleration. Overall, boolean descriptors exhibit significant gaps in accuracy and stability compared to floating-point descriptors, making them currently unsuitable for tasks that demand precision and reliability.

\begin{figure*}
    \centering
    \includegraphics[width=0.98\textwidth]{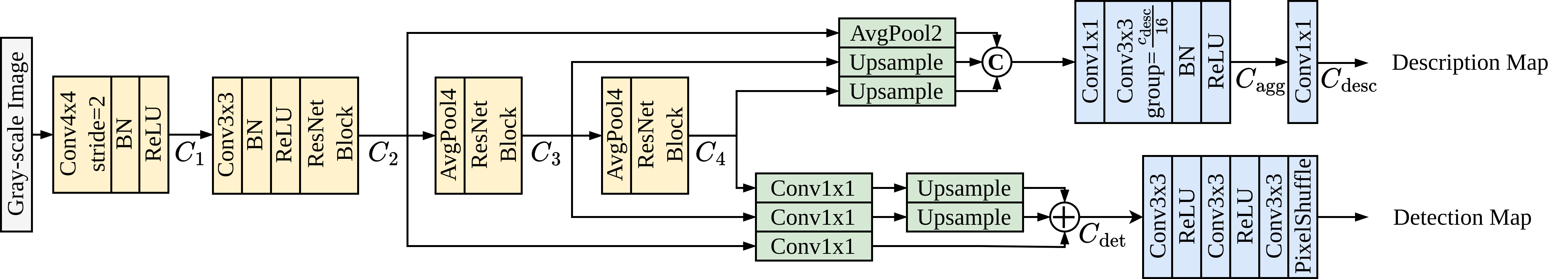}
    \caption{\raggedright \textbf{EdgePoint2 model architecture.} The feature encoder generates multi-scale features, as illustrated in the yellow block. These pyramid features are aggregated using various sizes and operations, as shown in the green block. The detection head and the description head, referred to as the blue block, utilize feature maps of sizes $\frac{H}{2} \times \frac{W}{2}$ and $\frac{H}{4} \times \frac{W}{4}$ to enhance accuracy and efficiency. During post-processing, keypoints are extracted from the detection map using non-maximum suppression (NMS), while the description map is sampled bilinearly. The notation $C_{i}, i\in\{1,2,3,4\}$ denotes the number of output channels for each block. $C_{\mathrm{agg}}$ and $C_{\mathrm{det}}$ denote the number of feature map channels for description and detection respectively and the $C_{\mathrm{desc}}$ is the descriptor dimension. The symbols $(\mathrm{\mathbf{C}})$ and $(\boldsymbol{+})$ represent the concatenation and addition operations, respectively.}
    \label{fig:network-architecture}
\end{figure*}

\subsection{Descriptor Distillation}

Descriptor distillation methodologies are still relatively limited in scope. Early approaches, such as in~\citep{HF-Net}, utilize mean squared error (MSE) to directly align descriptors generated by student models with those from teacher models. Although straightforward to implement, this method has inherent limitations. Unlike simple knowledge distillation~\citep{yang2022focal, yang2022masked, wang2024layer, wang2024relation}, descriptor distillation emphasizes maintaining the relative distribution of descriptors, often evaluated through distance matrices like cosine similarity matrices. Additionally, student descriptors frequently have a lower dimensionality than teacher descriptors. Consequently, descriptor distillation requires preserving the relative distribution structure within a lower-dimensional space, a challenge that cannot be adequately addressed by using a simple MSE approach alone.

To address these challenges, alternative methodologies have emerged that focus on learning the distribution of the distance matrix associated with the teacher descriptors~\citep{Similarity-Preserving, AWDesc, DescriptorDistillation}. These methods facilitate cross-dimensional distillation, enabling the student model to extract meaningful information from potentially different dimensional spaces. They primarily rely on the guidance provided by the distance matrix, neglecting the pre-allocated feature space of the teacher descriptors. While EdgePoint~\citep{EdgePoint} explored leveraging a pre-allocated feature space with PCA, the distance matrix of the teacher descriptors is not optimally preserved, nor is descriptor consistency maintained under varying illumination and viewpoint conditions. Thus, these approaches still fall short of fully utilizing the potential of the descriptor space. 

\section{Methodology}

In this section, we first introduce the network architecture of EdgePoint2. Subsequently, we outline the data preparation process. Then, we detail our proposed training loss function. Finally, we present an overview of the implementations of our methods, including the network configurations.

\subsection{Model Architecture}

The EdgePoint2 network is designed to be compact and efficient, optimizing inference efficiency without compromising accuracy, as illustrated in Fig. \ref{fig:network-architecture}. Given a grayscale image sized $H \times W$, the model employs a lightweight encoder to extract features at three distinct scales, subsequently detecting and describing features at these various scales to maximize both inference speed and accuracy.

\PAR{Feature Encoder}
The feature encoder transforms the input grayscale image into multi-scale features. Performing convolutions on the original size feature map is highly inefficient; therefore, it is essential to reduce the feature map size as quickly as possible. To initially reduce the size of the feature map, it begins with a $4 \times 4$ convolution with a stride of 2. This is followed by a $3 \times 3$ convolution and a ResNet block~\citep{ResNet} to produce a feature map at $\frac{H}{2} \times \frac{W}{2}$. The feature map is then downsampled using $4 \times 4$ average pooling and further processed through another ResNet block to generate feature maps with resolutions of $\frac{H}{8} \times \frac{W}{8}$ and $\frac{H}{32} \times \frac{W}{32}$. All convolutional layers in the encoder are followed by normalization layers. The ReLU activation function is employed throughout to facilitate rapid inference. The entire backbone comprises only two standard convolutions, two ResNet blocks, and pooling operations. This simple and efficient encoder effectively balances the receptive field size with a rapid reduction in feature map dimensions, achieving a maximum receptive field of $206 \times 206$ pixels.

\PAR{Detection Head}
Detection head in~\citep{SuperPoint,KP2D} utilizes a $\frac{H}{8} \times \frac{W}{8}$ feature map for efficient detection, although this may compromise overall performance. While detecting at the original image size enhances detection accuracy, it imposes significant computational and storage overhead~\citep{DISK,ALIKE,ALIKED}. To address this issue, we select $\frac{H}{2} \times \frac{W}{2}$ feature map size, achieving a favorable trade-off between accuracy and speed. The pyramid features go through three $1 \times 1$ convolutions to reduce the number of channels. Following this, the pyramid features are combined using addition (rather than concatenation) to increase processing speed, after which they are processed through three additional convolutional layers and reshuffled back to the original image dimensions.

\PAR{Description Head}
For the description head, utilizing a larger feature map size is beneficial for generating high-quality descriptors but adversely affects inference speed, particularly because the dimensionality of the feature maps needs to match that of the descriptors~\citep{DISK,ALIKE,ALIKED}. To achieve a balance between performance and latency, we implement a feature map of size $\frac{H}{4}\times\frac{W}{4}$. The pyramid features are first resized to this scale and then concatenated. We begin with a $1\times1$ convolution, followed by a group convolution in which every 16 channels form a distinct group, effectively reducing the number of parameters and improving efficiency. Finally, a convolutional layer generates the description map.

\subsection{Data Preparation}

\PAR{Image Preprocessing.}
For each image $ \mathcal{I} $ in the training dataset, we resize the image to $ H \times W $, denoting this resized image as $\mathcal{I}_1$. Subsequently, we apply data enhancements with color enhancement and random homography transformation to $\mathcal{I}_1$ for $N-1$ times, resulting in a mini set of images denoted as $ \mathcal{I}_2, \ldots, \mathcal{I}_N $. This process yields a set of $N$ images with $N \geq 2$. We adopt these synthetic data mentioned above. The detection and description cache are generated only using $\mathcal{I}_1$.

\PAR{Detection Cache.}
We utilize the ALIKED-N32~\citep{ALIKED} to generate the detection cache. We observe that the ALIKED exhibits a directional bias in keypoint detection, which may affect keypoint repeatability, particularly as the model size reduces. To overcome this challenge, we first generate the detection results using the original image $\mathcal{I}_1$. Then, we leverage the horizontally flipped image of $\mathcal{I}_1$, referred as $\overline{\mathcal{I}_1}$, to produce additional detection results, which are subsequently reverted to their original orientation. Finally, we apply non-maximum suppression (NMS) to merge the detection results, denoted as $\mathcal{Y}_\mathrm{t}$. This process is detailed in Algorithm~\ref{KeypointCache}. This method effectively mitigates the effects of directional bias and enhances the repeatability of keypoint detection.

\begin{algorithm}
    \caption{Generate Keypoint Cache}
    \label{KeypointCache}
    \KwIn{Image $\mathcal{I}_1$}
    \KwOut{Binary heatmap $\mathcal{Y}_\mathrm{t}$}

    \ \ 1: Generate keypoint $\boldsymbol{K}_1$ and its score $\boldsymbol{S}_1$ for $\mathcal{I}_1$.

    \ \ 2: Horizontally flip $\mathcal{I}_1$ to $\overline{\mathcal{I}_1}$.

    \ \ 3: Generate keypoint $\overline{\boldsymbol{K}_2}$ and its score $\boldsymbol{S}_2$ for $\overline{\mathcal{I}_1}$.

    \ \ 4: Horizontally flip $\overline{\boldsymbol{K}_2}$ to $\boldsymbol{K}_2$.

    \ \ 5: Let $\boldsymbol{K}_c = [\boldsymbol{K}_1; \boldsymbol{K}_2]$, $\boldsymbol{S}_c = [\boldsymbol{S}_1; \boldsymbol{S}_2]$.

    \ \ 6: Apply NMS on keypoint $\boldsymbol{K}_c$ with score $\boldsymbol{S}_c$ to obtain $\boldsymbol{K}_\mathrm{t}$.

    \ \ 7: Convert the keypoint $\boldsymbol{K}_\mathrm{t}$ into binary heatmap $\mathcal{Y}_\mathrm{t}$

    \textbf{Return: } $\mathcal{Y}_\mathrm{t}$
\end{algorithm}

\PAR{Description Cache.}
We cache 512 descriptors of 128 dimensions based on $\mathcal{I}_1$ using the ALIKE-L~\citep{ALIKE}. During the training process, we select the top-$C_{\mathrm{desc}}$ descriptors for $C_{\mathrm{desc}}$-dimensional student descriptors, under the condition that the number of the co-visible descriptors in $\mathcal{I}_1, \ldots, \mathcal{I}_N$ are at least $C_{\mathrm{desc}}$. If the number of co-visible descriptors is less than $C_{\mathrm{desc}}$, the corresponding set of images is discarded. The selected top-$C_{\mathrm{desc}}$ descriptors are denoted as $\boldsymbol{D}_\mathrm{t}\in \mathbb{R}^{C_{\mathrm{desc}} \times 128}$.

\subsection{Descriptor Loss}

\subsubsection{Orthogonal Procrustes Loss}

We follow the general definition for local descriptors, which states that the descriptors are L2-normalized, meaning that their values are scaled to have a unit Euclidean norm. Since the distance between two unit-length descriptors can be represented by their cosine distance, descriptor distillation aims to preserve the cosine distance matrix among keypoints, i.e,
\begin{equation}
\begin{aligned}
    \operatorname*{min}_{\boldsymbol{D}_{\mathrm{s},i}} & \ \| \boldsymbol{D}_\mathrm{t} \boldsymbol{D}_\mathrm{t}^\mathrm{T} - \boldsymbol{D}_{\mathrm{s},i} \boldsymbol{D}_{\mathrm{s},i}^\mathrm{T} \|_\mathrm{F}^2, \\
\end{aligned}
\label{DistillationTarget}
\end{equation}
where $\boldsymbol{D}_{\mathrm{s},i}\in\mathbb{R}^{C_{\mathrm{desc}} \times C_{\mathrm{desc}}}$ denote the student descriptors sampled from image $\mathcal{I}_i$, $i \in \{1, \ldots, N\}$. Typically, in a distillation task, the dimension of the student descriptor is smaller than that of the teacher descriptor, which implies that $C_{\mathrm{desc}} < 128$. Hence, the rank of $\boldsymbol{D}_\mathrm{t} \boldsymbol{D}_\mathrm{t}^\mathrm{T}$ is no less than $\boldsymbol{D}_{\mathrm{s},i} \boldsymbol{D}_{\mathrm{s},i}^\mathrm{T}$. As a result, we can consider this problem as a Low Rank Approximation (LRA), where the optimal solution can be obtained through Singular Value Decomposition (SVD). Furthermore, since $\boldsymbol{D}_\mathrm{t} \boldsymbol{D}_\mathrm{t}^\mathrm{T}$ is symmetric, the SVD resolution for $\boldsymbol{D}_\mathrm{t} \boldsymbol{D}_\mathrm{t}^\mathrm{T}$ can be directly computed from $\boldsymbol{D}_\mathrm{t}$ as follows
\begin{equation}
\begin{aligned}
    \boldsymbol{D}_\mathrm{t} &= \boldsymbol{U}_\mathrm{t}\boldsymbol{\Sigma}_\mathrm{t}\boldsymbol{V}_\mathrm{t}^\mathrm{T}, \\
    \boldsymbol{D}_\mathrm{l} &= \begin{bmatrix} 
        \boldsymbol{\sigma}_{\mathrm{t},1} \boldsymbol{u}_{\mathrm{t},1}, ~
        \boldsymbol{\sigma}_{\mathrm{t},2} \boldsymbol{u}_{\mathrm{t},2}, ~
        \cdots ~ 
        \boldsymbol{\sigma}_{\mathrm{t},C_{\mathrm{desc}}} \boldsymbol{u}_{\mathrm{t},C_{\mathrm{desc}}} 
        \end{bmatrix}, \\
\end{aligned}
\end{equation}
where $\boldsymbol{U}_\mathrm{t}\boldsymbol{\Sigma}_\mathrm{t}\boldsymbol{V}_\mathrm{t}^\mathrm{T}$ is the SVD of $\boldsymbol{D}_\mathrm{t}$, $\boldsymbol{u}_{\mathrm{t},i}$ is the $i$-th column of $\boldsymbol{U}_\mathrm{t}$, $\boldsymbol{\sigma}_{\mathrm{t},1}$ is the $i$-th singular value of $\boldsymbol{D}_\mathrm{t}$, and $\boldsymbol{D}_\mathrm{l}$ represents the compressed descriptors. Since the number of selected descriptors corresponds to the desired dimension $C_{\mathrm{desc}}$, and $C_{\mathrm{desc}}$ is smaller than the dimension of the teacher descriptor, this procedure is guaranteed to be lossless, meaning that $\boldsymbol{D}_\mathrm{l}\boldsymbol{D}_\mathrm{l}^\mathrm{T} = \boldsymbol{D}_\mathrm{t}\boldsymbol{D}_\mathrm{t}^\mathrm{T}$. 

\begin{figure*}[htbp]
    \centering
    \includegraphics[width=0.85\textwidth]{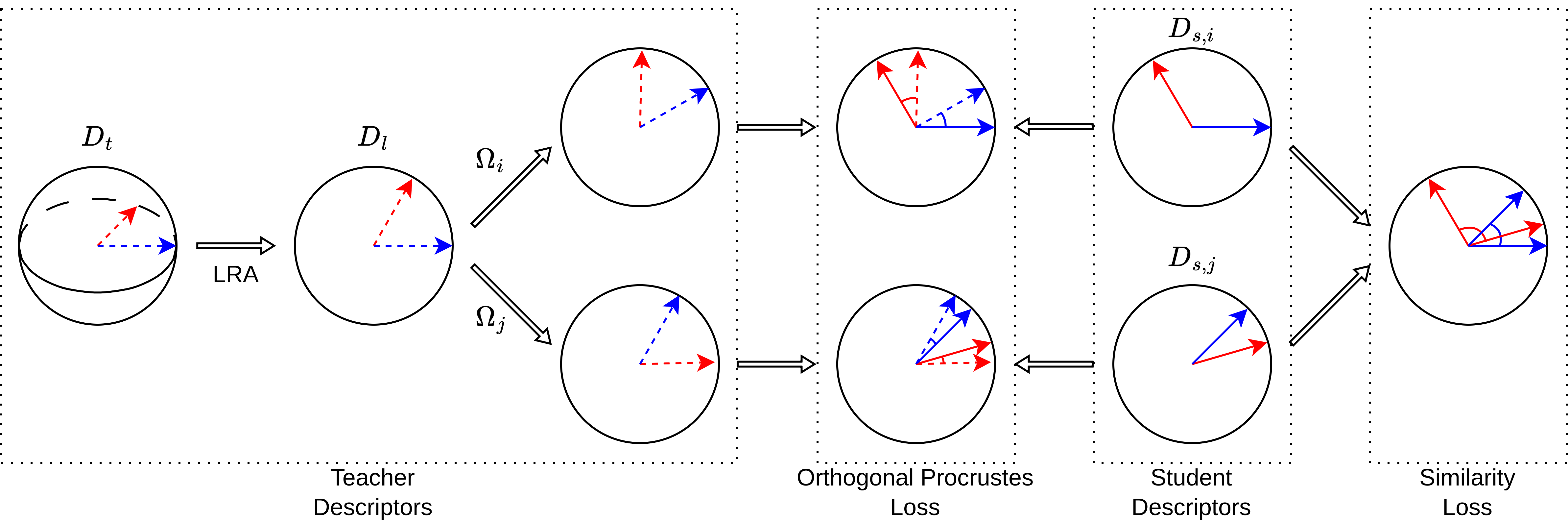}
    \caption{\raggedright \textbf{Illustration of Orthogonal Procrustes loss and similarity loss.} The dashed and solid lines represent the descriptors of the teacher and student, respectively. The red and blue arrows indicate descriptors sampled from different locations. For illustration, the teacher descriptors, denoted as $\boldsymbol{D}_\mathrm{t}$, are initially represented within a three-dimensional unit sphere and are subsequently compressed to two dimensions using LRA, referred to as $\boldsymbol{D}_\mathrm{l}$. Given the student descriptors $\boldsymbol{D}_{\mathrm{s},i}$ and $\boldsymbol{D}_{\mathrm{s},j}$, where $i, j \in \{1,2,...,N\}, i \neq j$, the orthogonal matrices $\boldsymbol{\Omega}_i$ and $\boldsymbol{\Omega}_j$ can be determined through Eq.~\eqref{opp_solution}. Subsequently, we can compute the Orthogonal Procrustes loss, which quantifies the cosine distances between the designated descriptors. For the similarity loss, we compute the cosine distance for each descriptor within mini set. The Orthogonal Procrustes loss emphasizes the relative position in embedding space between the descriptors extracted from the same image, while the similarity loss ensures that the corresponding descriptors from different images remain sufficiently close.}
    \label{fig:descriptor-loss}
\end{figure*}

Now we obtain $\boldsymbol{D}_\mathrm{l}$, which is one of the optimal solutions for $\boldsymbol{D}_{\mathrm{s},i}$ in Eq.~\eqref{DistillationTarget}. However, since $\boldsymbol{D}_\mathrm{l}$ is computed for each mini set of teacher descriptors, the compression results vary when changing the mini set. Meanwhile, for any orthogonal matrix $\boldsymbol{\Omega}$ that satisfies $\boldsymbol{\Omega}\boldsymbol{\Omega}^\mathrm{T} = \boldsymbol{I}$, we have
\begin{equation}
\begin{aligned}
    (\boldsymbol{D}_\mathrm{l}\boldsymbol{\Omega})(\boldsymbol{D}_\mathrm{l}\boldsymbol{\Omega})^\mathrm{T} = \boldsymbol{D}_\mathrm{l}\boldsymbol{\Omega}\boldsymbol{\Omega}^\mathrm{T}\boldsymbol{D}_\mathrm{l}^\mathrm{T} = \boldsymbol{D}_\mathrm{l}\boldsymbol{D}_\mathrm{l}^\mathrm{T}. \\
\end{aligned}
\end{equation}
Thus we can define our Orthogonal Procrustes loss as
\begin{equation}
\begin{aligned}
    L_{\mathrm{op}} &= \frac{1}{N} \sum_{i=1}^{N} \| \boldsymbol{D}_\mathrm{l} \boldsymbol{\Omega}_i - \boldsymbol{D}_{\mathrm{s},i} \|_\mathrm{F}^2, \\
\end{aligned}
\label{loss_desc_op_v1}
\end{equation}
and $\boldsymbol{\Omega}_i$ is the solution for Orthogonal-Procrustes Problem~\citep{OPP}
\begin{equation}
\centering
\label{opp_solution}
\begin{aligned}
    \boldsymbol{\Omega}_i &= \operatorname*{arg\,min}_{\boldsymbol{X} \in \mathbb{O}(C_{\mathrm{desc}},\mathbb{R})} \ \|\boldsymbol{D}_\mathrm{l} \boldsymbol{X} - \boldsymbol{D}_{\mathrm{s},i}\|_\mathrm{F}^2 \\
             &= \operatorname*{arg\,min}_{\boldsymbol{X} \in \mathbb{O}(C_{\mathrm{desc}},\mathbb{R})} \ \|\boldsymbol{D}_\mathrm{l} \boldsymbol{X}\|_\mathrm{F}^2 + \|\boldsymbol{D}_{\mathrm{s},i}\|_\mathrm{F}^2 - 2\mathrm{tr}(\boldsymbol{D}_\mathrm{l}\boldsymbol{X}\boldsymbol{D}_{\mathrm{s},i}^\mathrm{T}) \\
             &= \operatorname*{arg\,min}_{\boldsymbol{X} \in \mathbb{O}(C_{\mathrm{desc}},\mathbb{R})} \ \underbrace{\|\boldsymbol{D}_\mathrm{l}\|_\mathrm{F}^2 + \|\boldsymbol{D}_{\mathrm{s},i}\|_\mathrm{F}^2}_{\text{constant}} - 2\mathrm{tr}(\boldsymbol{D}_\mathrm{l}\boldsymbol{X}\boldsymbol{D}_{\mathrm{s},i}^\mathrm{T}) \\
             &= \operatorname*{arg\,max}_{\boldsymbol{X} \in \mathbb{O}(C_{\mathrm{desc}},\mathbb{R})} \ \mathrm{tr}(\boldsymbol{D}_\mathrm{l}\boldsymbol{X}\boldsymbol{D}_{\mathrm{s},i}^\mathrm{T}) \\
             &= \operatorname*{arg\,max}_{\boldsymbol{X} \in \mathbb{O}(C_{\mathrm{desc}},\mathbb{R})} \ \mathrm{tr}(\boldsymbol{D}_{\mathrm{s},i}^\mathrm{T} \boldsymbol{D}_\mathrm{l} \boldsymbol{X}) \\
             &= \operatorname*{arg\,max}_{\boldsymbol{X} \in \mathbb{O}(C_{\mathrm{desc}},\mathbb{R})} \ \mathrm{tr}(\boldsymbol{U}_i \boldsymbol{\Sigma}_i \boldsymbol{V}_i^\mathrm{T} \boldsymbol{X}) \\
             &= \boldsymbol{V}_i \boldsymbol{U}_i^\mathrm{T},
\end{aligned}
\end{equation}
where $\mathbb{O}(C_{\mathrm{desc}},\mathbb{R})$ denotes the orthogonal group, $\boldsymbol{U}_i \boldsymbol{\Sigma}_i \boldsymbol{V}_i^\mathrm{T}$ is the SVD of $\boldsymbol{D}_{\mathrm{s},i}^\mathrm{T} \boldsymbol{D}_\mathrm{l}$.

In practice, we first use Eq.~\eqref{opp_solution} to obtain $\boldsymbol{\Omega}_i$ to minimize the Orthogonal Procrustes loss, then we apply the back propagation optimizer.

\subsubsection{Similarity Loss}

The Orthogonal Procrustes loss aims to preserve the cosine distance matrix from Eq.~\eqref{DistillationTarget}. Given the set of images $\mathcal{I}_i$ with data enhancements, we also need to minimize the distance between the sampled descriptors corresponding to the same pixels within each enhanced image set. Thus, using the sampled descriptors $\boldsymbol{D}_{\mathrm{s},i}$, we can define the similarity loss as
\begin{equation}
L_{\mathrm{sim}} = \frac{1}{N(N-1)} \sum_{i=1}^{N} \sum_{j=i+1}^{N} \| \boldsymbol{D}_{\mathrm{s},i} - \boldsymbol{D}_{\mathrm{s},j} \|_\mathrm{F}^2.
\label{loss_similarity}
\end{equation}

\subsubsection{Descriptor Loss}

Our descriptor loss is a combination of Orthogonal Procrustes loss and similarity loss. Given a batch of teacher descriptors for distillation, the Orthogonal Procrustes loss serves as an allocation method for the student descriptors to approximate the cosine distance matrix of the teacher descriptors. The similarity loss ensure that the student descriptors remain invariant to illumination and viewpoint change. The procedure of these losses are shown in Fig.~\ref{fig:descriptor-loss}. This approach allows for distilling teacher descriptors to \emph{any} dimension while simultaneously learning relative positional relationships in the embedding space. 

\subsection{Detection Loss}

We adopt the UnfoldSoftmax from our previous work~\citep{EdgePoint} for detection as
\begin{equation}
L_{\mathrm{detect}} = \sum_{p}^{\mathcal{P}} \operatorname{UnfoldSoftmax}\left(\boldsymbol{x}_{\mathrm{s},p}; \boldsymbol{y}_{\mathrm{t},p}\right),
\label{proposed_detect_loss}
\end{equation}
where $\mathcal{P}$ represents all the patches that can be extracted from a given image, and $\boldsymbol{x}_{\mathrm{s},p}$ and $\boldsymbol{y}_{\mathrm{t},p}$ are the corresponding vectors unfolded with kernel size $k\times k$ from the raw logits $\mathcal{X}_\mathrm{s}$ and the binary heatmap $\mathcal{Y}_\mathrm{t}$ respectively. During computation, we append a 0 to $\boldsymbol{x}_{\mathrm{s},p}$ and append {0,1} to $\boldsymbol{y}_{\mathrm{t},p}$ to indicate the absence or presence of an interest point in the patch. To reduce memory consumption and improve numerical stability, we implement the UnfoldSoftmax loss using only two convolution layers, as outlined in Algorithm~\ref{UnfoldSoftmax}. In practice, we choose a kernel size $ k = 5 $ for the detection loss.

\begin{algorithm}[t]
    \caption{UnfoldSoftmax loss using PyTorch}
    \label{UnfoldSoftmax}
    \KwIn{Raw logits $\mathcal{X}_\mathrm{s}$, binary heatmap $\mathcal{Y}_\mathrm{t}$, kernel size $k$}
    \KwOut{Loss $L_{\mathrm{detect}}$}

    \ \ 1: Initialize $\mathcal{W} = \mathrm{torch.ones}(1,1,k,k)$.

    \ \ 2: $l_1=\mathrm{F.conv2d}(\mathcal{X}_\mathrm{s} \cdot \mathcal{Y}_\mathrm{t}, \mathcal{W})$.

    \ \ 3: $l_2=\mathrm{F.conv2d}(\mathcal{X}_\mathrm{s}.\mathrm{exp}(), \mathcal{W})+1$.

    \ \ 4: $L_{\mathrm{detect}}= - (l_1 - l_2.\mathrm{log}()).\mathrm{mean}()$.

    \textbf{Return: } $L_{\mathrm{detect}}$
\end{algorithm}

\subsection{Implementation Details}

The overall loss function is defined as a weighted sum of the individual losses:
\begin{equation}
\centering
L_{\mathrm{total}} = w_{\mathrm{op}} L_{\mathrm{op}} + w_{\mathrm{sim}} L_{\mathrm{sim}} + w_{\mathrm{detect}} L_{\mathrm{detect}},
\end{equation}
where $w_{\mathrm{desc}}=0.5$, $w_{\mathrm{sim}}=0.1$ and $w_{\mathrm{detect}}=1$ in practice.

To satisfy diverse application requirements, we offer 14 sub-models, as outlined in Table \ref{tab:network-configurations}. All sub-models are trained on Megadepth~\citep{MegaDepth} and COCO~\citep{COCO} datasets. Each image is resized to 512 $\times$ 512 pixels, and color enhancement and random homography transformation are applied to generate an image set comprising $N=4$ images.

Training is conducted across two GPUs for three epochs, with a batch size of 64 per GPU, resulting in a total batch size of 128. Our largest model, EdgePoint2-E64, requires no more than 24 GB of memory on each GPU. We employ the AdamW optimizer~\citep{AdamW}, beginning with an initial learning rate of 0.002, which is halved after each epoch. Due to the computational demands of SVD, training times range from 3 to 4 hours for models with descriptor dimensions of 32 and 64.

\begin{table}[t]
\centering
\caption{\textbf{EdgePoint2 network configurations.} We denote the models as EdgePoint2-[T/S/M/L/E][32/48/64] to indicate the model size and descriptor dimension in the experiments.}
\resizebox{\columnwidth}{!}{
\begin{tabular}{cccccccc}
\hline
Models & $C_1$ & $C_2$ & $C_3$ & $C_4$ & $C_{\mathrm{agg}}$ & \multicolumn{1}{l}{$C_{\mathrm{det}}$} & \multicolumn{1}{l}{$C_{\mathrm{desc}}$} \\ \hline
Tiny            & 8     & 8     & 16    & 24    & 48                 & 8                                      & 32/48                                   \\
Small           & 8     & 8     & 24    & 32    & 64                 & 8                                      & 32/48/64                                \\
Medium          & 8     & 16    & 32    & 48    & 96                 & 8                                      & 32/48/64                                \\
Large           & 8     & 16    & 48    & 64    & 128                & 8                                      & 32/48/64                                \\
Enormous        & 16    & 16    & 48    & 64    & 128                & 16                                     & 32/48/64                                \\ \hline
\end{tabular}
}
\label{tab:network-configurations}
\end{table}

\section{Experiments}

This section begins with a comparison of the computational resources required by our method with those of SOTA methods on embedded devices, on both CPU and GPU. Subsequently, we evaluate the performance of our method against SOTA methods across various tasks and benchmarks, including homography estimation, relative pose estimation, FM-Bench, and visual localization. Finally, we present an ablation study to demonstrate the effectiveness of our descriptor loss.

For comparison, we assess several SOTA methods, including SuperPoint~\citep{SuperPoint}, DISK~\citep{DISK}, ALIKE-[T/S/N/L]~\citep{ALIKE}, ALIKED-[T16/N16/N32]~\citep{ALIKED}, AWDesc-[T16/T32/CA]~\citep{AWDesc}, XFeat~\citep{XFeat}, and EdgePoint~\citep{EdgePoint}. During inference, NMS is applied with a size of 2 and a detection threshold of -5 for EdgePoint2. By default, we extract a total of 4096 keypoints for each image for the experiments, unless specified otherwise.

\subsection{Runtime Efficiency}

\PAR{Setup}
We evaluate the descriptor dimension (Dim), the number of millions of parameters (MP), and the GFLOPs required by these models. Additionally, we assess the model runtime in two common environments: edge devices equipped with an accelerator and those relying solely on CPU. The NVIDIA Jetson Orin-NX, a widely used edge embedded device, features an 8-core ARM CPU and an integrated GPU. We execute all the networks on Orin-NX GPU using TensorRT and on the Orin-NX CPU with OpenVINO, both of which are standard libraries for optimizing and deploying deep learning models in edge applications. The runtime comparisons are based on frames per second (FPS) with an input size of $480 \times 640$ and extracting 1024 keypoints.

\begin{table*}
\centering
\caption{\raggedright \textbf{Computation Resources Comparison.} The top three best results are highlighted in \textcolor{cr}{red}, \textcolor{cg}{green} and \textcolor{cb}{blue}. * denotes running native PyTorch on CPU due to OpenVINO runtime failure.}
\begin{tabular}{c@{\hskip 5pt}lccccc}
\hline
\multicolumn{1}{l}{\multirow{2}{*}{}} & \multicolumn{1}{c}{\multirow{2}{*}{\textbf{Method}}} & \multirow{2}{*}{\textbf{Dim}} & \multirow{2}{*}{\textbf{MP}} & \multirow{2}{*}{\textbf{GFLOPs}} & \multicolumn{2}{c}{\textbf{FPS on Orin-NX}} \\
\multicolumn{1}{l}{}                  & \multicolumn{1}{c}{}                                 &                               &                              &                                  & GPU                & CPU  \\ \hline
\multirow{12}{*}{\rotatebox{90}{Standard}}            & SuperPoint~\citep{SuperPoint}                                           & 256                           & 1.301                        & 26.11                            & 124.2                       & 3.32          \\
                                      & DISK~\citep{DISK}                                                 & 128                           & 1.092                        & 98.97                            & 26.9                        & 0.54$^*$      \\
                                      & ALIKE-T~\citep{ALIKE}                                              & 64                            & 0.080                       & 2.11                             & 126.6                      & 9.22          \\
                                      & ALIKE-S~\citep{ALIKE}                                              & 96                            & 0.142                       & 3.89                             & 94.8                       & 6.46          \\
                                      & ALIKE-N~\citep{ALIKE}                                              & 128                           & 0.318                        & 7.91                             & 88.3                      & 4.41          \\
                                      & ALIKE-L~\citep{ALIKE}                                              & 128                           & 0.653                        & 19.68                            & 70.4                        & 2.63          \\
                                      & ALIKED-T16~\citep{ALIKED}                                           & 64                            & 0.192                       & 1.37                             & 52.5   & 1.59$^*$      \\
                                      & ALIKED-N16~\citep{ALIKED}                                           & 128                           & 0.677                        & 4.05                             & 31.4   & 1.12$^*$      \\
                                      & ALIKED-N32~\citep{ALIKED}                                           & 128                           & 0.980                        & 4.62                             & 29.7   & 1.12$^*$      \\
                                      & AWDesc-T16~\citep{AWDesc}                                           & 128                           & 0.172                        & 4.50                             & 180.4                      & 2.22$^*$      \\
                                      & AWDesc-T32~\citep{AWDesc}                                           & 128                           & 0.390                        & 11.81                            & 168.2                       & 1.34$^*$      \\
                                      & AWDesc-CA~\citep{AWDesc}                                            & 128                           & 10.13                        & 27.45                            & 84.0                        & 2.49          \\ \hdashline
\multirow{16}{*}{\rotatebox{90}{Fast}}                & XFeat~\citep{XFeat}                                                & 64                            & 0.658                        & 1.31                             & 277.4                       & 35.33         \\
                                      & EdgePoint~\citep{EdgePoint}                                            & 32                            & \textcolor{cb}{\textbf{0.030}}                        & \textcolor{cr}{\textbf{0.36}}                             & \textcolor{cr}{\textbf{838.8}}                       & \textcolor{cr}{\textbf{80.78}}         \\
                                      & EdgePoint2-T32                                       & 32                            & \textcolor{cr}{\textbf{0.028}}                        & \textcolor{cg}{\textbf{0.49}}                             & 508.7                       & \textcolor{cg}{\textbf{40.84}}         \\
                                      & EdgePoint2-T48                                       & 48                            & \textcolor{cr}{\textbf{0.028}}                        & \textcolor{cb}{\textbf{0.50}}                             & 503.0                       & \textcolor{cb}{\textbf{36.41}}         \\
                                      & EdgePoint2-S32                                       & 32                            & 0.044                        & 0.60                             & \textcolor{cg}{\textbf{537.2}}                       & 36.52         \\
                                      & EdgePoint2-S48                                       & 48                            & 0.045                        & 0.62                             & \textcolor{cb}{\textbf{525.3}}                       & 34.05         \\
                                      & EdgePoint2-S64                                       & 64                            & 0.046                        & 0.64                             & 522.3                       & 33.42         \\
                                      & EdgePoint2-M32                                       & 32                            & 0.086                        & 1.16                             & 454.1                       & 27.28         \\
                                      & EdgePoint2-M48                                       & 48                            & 0.087                        & 1.19                             & 448.4                       & 26.13         \\
                                      & EdgePoint2-M64                                       & 64                            & 0.089                        & 1.22                             & 437.2                       & 25.04         \\
                                      & EdgePoint2-L32                                       & 32                            & 0.144                        & 1.48                             & 431.2                       & 23.26         \\
                                      & EdgePoint2-L48                                       & 48                            & 0.146                        & 1.52                             & 421.2                       & 22.07         \\
                                      & EdgePoint2-L64                                       & 64                            & 0.149                        & 1.56                             & 415.3                       & 21.22         \\
                                      & EdgePoint2-E32                                       & 32                            & 0.151                        & 1.88                             & 386.4                       & 17.74         \\
                                      & EdgePoint2-E48                                       & 48                            & 0.153                        & 1.92                             & 379.1                       & 17.06         \\
                                      & EdgePoint2-E64                                       & 64                            & 0.155                        & 1.96                             & 375.7                       & 16.92         \\ \hline
\end{tabular}
\label{tab:system-runtime}
\end{table*}

\begin{figure}
    \centering
    \includegraphics[trim=0 0 0 0, clip, width=0.45\textwidth]{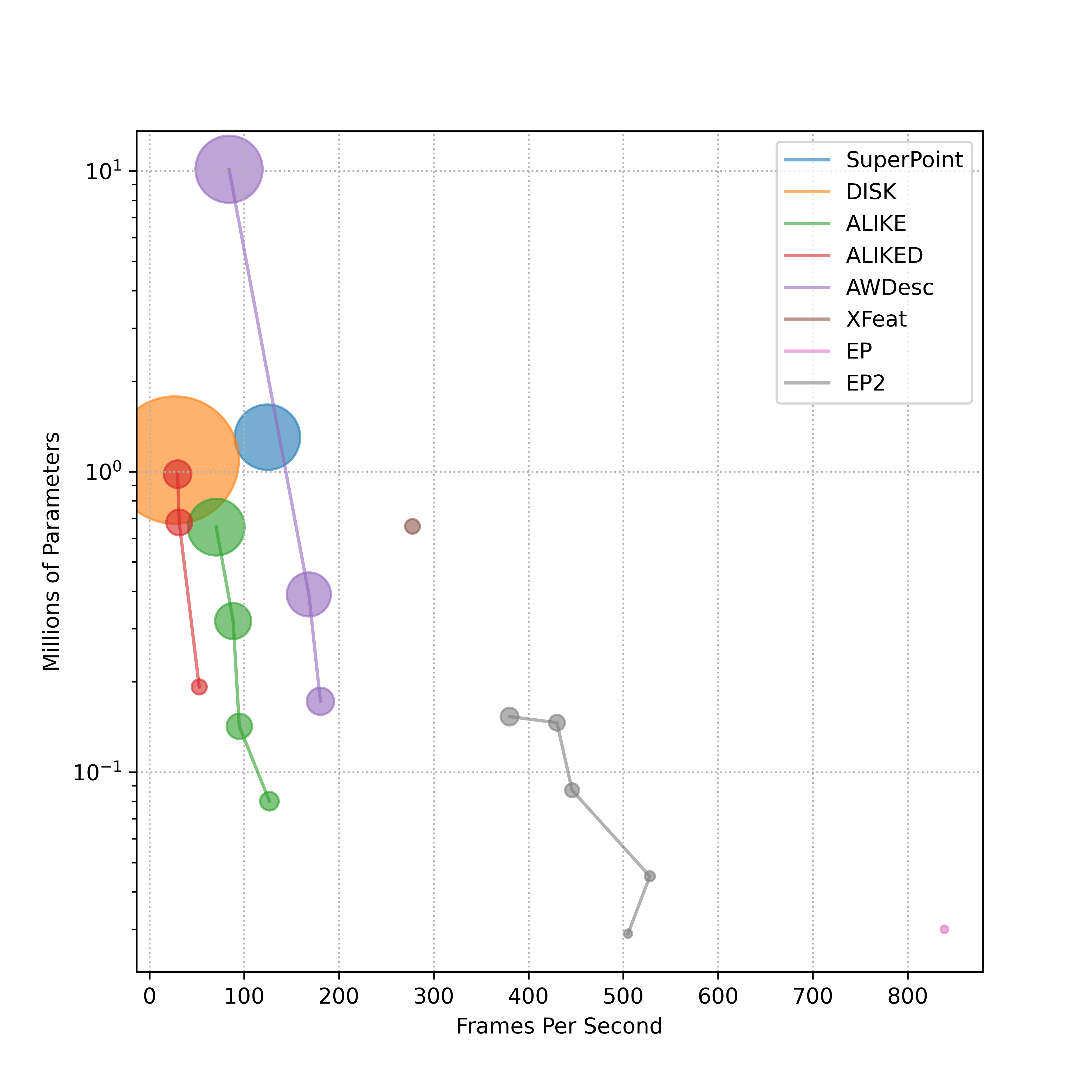}
    \caption{\raggedright \textbf{Runtime efficiency comparison.} The x-axis represents FPS on GPU, the y-axis denotes the number of parameters in log scale, and the size of the circles indicates the relative computational cost.}
    \label{fig:runtime}
\end{figure}

\PAR{Results}
As shown in Table \ref{tab:system-runtime}, our EdgePoint2 32-dim models achieve the smallest descriptor dimension, while the 64-dim models do not exceed the size of any SOTA models. The EdgePoint2-T models utilize the fewest parameters (28k) and require the least computational resources (0.49 GFLOPs). On Orin-NX GPU, all EdgePoint2 series models are slightly slower than our previous model EdgePoint but outperform all other models. Our largest model EdgePoint2-E is a lot faster than ALIKED-n32 (12$\times$) and AWDesc-CA (4.5 $\times$). On ARM CPU, our EdgePoint2-[T/S/M] models achieve real-time inference. As illustrated in Fig.~\ref{fig:runtime}, the EdgePoint2 series utilizes significantly fewer parameters and computational resources. Even when compared to models with an equivalent number of parameters or computational complexity, EdgePoint2 exhibits a remarkably higher inference speed. This highlights the efficiency of our network architecture design for deployment on embedded devices.


\subsection{Homography Estimation}

\PAR{Setup}
The HPatches~\citep{HPatches} dataset comprises 116 scenes, each containing six images, and specifically designed to evaluate descriptors under varying illumination and viewpoint changes. Following the procedure outlined in~\citep{ALIKE,D2-Net}, we exclude eight unreliable scenes and employ a mutual nearest neighbor approach as the matcher. MAGSAC++~\citep{MAGSAC} is utilized for robust homography estimation. For evaluation metrics, we adopt the Mean Homography Accuracy (MHA) under thresholds of {1,3,5} pixels for illumination and viewpoint changes.

\begin{table*}
\centering
\caption{\raggedright \textbf{Homography estimation on HPatches.} The top three best results in each sector are highlighted in \textcolor{cr}{red}, \textcolor{cg}{green} and \textcolor{cb}{blue}.}
\begin{tabular}{clcccccc}
\hline
\multirow{2}{*}{}               & \multicolumn{1}{c}{\multirow{2}{*}{\textbf{Method}}} & \multicolumn{3}{c}{\textbf{MHA-Illumination}}                            & \multicolumn{3}{c}{\textbf{MHA-ViewPoint}}                               \\
                                & \multicolumn{1}{c}{}                                 & \multicolumn{1}{c}{@1} & \multicolumn{1}{c}{@3} & \multicolumn{1}{c}{@5} & \multicolumn{1}{c}{@1} & \multicolumn{1}{c}{@3} & \multicolumn{1}{c}{@5} \\ \hline
\multirow{9}{*}{\rotatebox{90}{\textbf{96 / 128 / 256 dim}}} & SuperPoint~\citep{SuperPoint}                          & 63.46                  & 94.23                  & 98.46                  & 37.14                  & 69.64                  & 80.00                  \\
                                & DISK~\citep{DISK}                                      & \textcolor{cr}{\textbf{68.46}}         & \textcolor{cb}{\textbf{95.00}}         & 98.46                  & 35.36                  & 65.00                  & 78.93                  \\
                                & ALIKE-S~\citep{ALIKE}                                  & 63.08                  & 93.46                  & 98.46                  & 31.43                  & 70.00                  & 78.93                  \\
                                & ALIKE-N~\citep{ALIKE}                                  & \textcolor{cb}{\textbf{66.54}}         & 94.23                  & 98.46                  & 38.21                  & 69.29                  & 79.64                  \\
                                & ALIKE-L~\citep{ALIKE}                                  & 65.00                  & 94.62                  & \textcolor{cr}{\textbf{99.23}}         & 39.29                  & 70.00                  & 81.79                  \\
                                & ALIKED-N16~\citep{ALIKED}                              & 62.69                  & \textcolor{cb}{\textbf{95.38}}         & \textcolor{cg}{\textbf{98.85}}         & \textcolor{cb}{\textbf{41.43}}         & \textcolor{cb}{\textbf{73.21}}         & 82.50                  \\
                                & ALIKED-N32~\citep{ALIKED}                              & 61.54                  & 94.62                  & \textcolor{cg}{\textbf{98.85}}         & 40.00                  & \textcolor{cb}{\textbf{73.21}}         & \textcolor{cg}{\textbf{83.21}}         \\
                                & AWDesc-T32~\citep{AWDesc}                              & \textcolor{cg}{\textbf{67.31}}         & \textcolor{cr}{\textbf{95.77}}         & 98.46                  & \textcolor{cg}{\textbf{46.79}}         & \textcolor{cg}{\textbf{75.36}}         & \textcolor{cb}{\textbf{82.86}}         \\
                                & AWDesc-CA~\citep{AWDesc}                               & 61.15                  & 94.23                  & \textcolor{cg}{\textbf{98.85}}         & \textcolor{cr}{\textbf{47.50}}         & \textcolor{cr}{\textbf{76.43}}         & \textcolor{cr}{\textbf{86.43}}         \\ \hdashline
\multirow{8}{*}{\rotatebox{90}{\textbf{64 dim}}}         & ALIKE-T~\citep{ALIKE}                                  & 65.77                  & 94.23                  & 98.46                  & 32.14                  & 68.21                  & 76.79                  \\
                                & ALIKED-T16~\citep{ALIKED}                              & 60.77                  & \textcolor{cr}{\textbf{95.00}}         & \textcolor{cr}{\textbf{98.85}}         & \textcolor{cb}{\textbf{41.43}}         & \textcolor{cr}{\textbf{73.57}}         & \textcolor{cg}{\textbf{82.86}}         \\
                                & AWDesc-T16~\citep{AWDesc}                              & 63.08                  & 94.23                  & 98.08                  & \textcolor{cr}{\textbf{45.36}}         & \textcolor{cg}{\textbf{73.21}}         & \textcolor{cr}{\textbf{83.57}}         \\
                                & XFeat~\citep{XFeat}                                    & 62.31                  & \textcolor{cg}{\textbf{94.62}}         & 97.69                  & 32.86                  & 66.79                  & 81.07                  \\
                                & EdgePoint2-S64                                       & 63.85                  & 93.85                  & 98.08                  & 38.21                  & 72.50                  & 81.07                  \\
                                & EdgePoint2-M64                                       & \textcolor{cr}{\textbf{67.31}}         & \textcolor{cg}{\textbf{94.62}}         & \textcolor{cr}{\textbf{98.85}}         & 41.07                  & 72.14                  & 81.07                  \\
                                & EdgePoint2-L64                                       & \textcolor{cg}{\textbf{66.92}}         & \textcolor{cg}{\textbf{94.62}}         & \textcolor{cr}{\textbf{98.85}}         & \textcolor{cg}{\textbf{43.57}}         & 71.43                  & 80.71                  \\
                                & EdgePoint2-E64                                       & \textcolor{cb}{\textbf{66.15}}         & 93.46                  & 98.46                  & \textcolor{cb}{\textbf{41.43}}         & \textcolor{cb}{\textbf{72.86}}         & \textcolor{cg}{\textbf{83.21}}         \\ \hdashline
\multirow{5}{*}{\rotatebox{90}{\textbf{48 dim}}}         & EdgePoint2-T48                                       & 65.38                  & \textcolor{cr}{\textbf{94.23}}         & 98.08                  & 36.07                  & 69.64                  & 80.36                  \\
                                & EdgePoint2-S48                                       & \textcolor{cr}{\textbf{68.08}}         & 93.85                  & 98.46                  & 37.50                  & \textcolor{cg}{\textbf{73.21}}         & 79.64                  \\
                                & EdgePoint2-M48                                       & \textcolor{cb}{\textbf{66.15}}         & \textcolor{cr}{\textbf{94.23}}         & \textcolor{cg}{\textbf{98.85}}         & \textcolor{cg}{\textbf{40.71}}         & \textcolor{cb}{\textbf{72.14}}         & \textcolor{cr}{\textbf{83.21}}         \\
                                & EdgePoint2-L48                                       & \textcolor{cg}{\textbf{66.54}}         & \textcolor{cr}{\textbf{94.23}}         & \textcolor{cg}{\textbf{98.85}}         & \textcolor{cr}{\textbf{41.07}}         & 71.79                  & \textcolor{cb}{\textbf{81.79}}         \\
                                & EdgePoint2-E48                                       & 65.77                  & \textcolor{cr}{\textbf{94.23}}         & \textcolor{cr}{\textbf{99.23}}         & \textcolor{cg}{\textbf{40.71}}         & \textcolor{cr}{\textbf{73.93}}         & \textcolor{cr}{\textbf{83.21}}         \\ \hdashline
\multirow{6}{*}{\rotatebox{90}{\textbf{32 dim}}}         & EdgePoint~\citep{EdgePoint}                            & 63.46                  & \textcolor{cb}{\textbf{93.85}}         & 98.08                  & 33.57                  & 65.00                  & 76.79                  \\
                                & EdgePoint2-T32                                       & \textcolor{cb}{\textbf{64.62}}         & 93.08                  & \textcolor{cg}{\textbf{98.46}}         & 37.14                  & 69.64                  & 80.71                  \\
                                & EdgePoint2-S32                                       & 63.46                  & \textcolor{cr}{\textbf{94.23}}         & \textcolor{cg}{\textbf{98.46}}         & \textcolor{cb}{\textbf{37.50}}         & 71.07                  & 80.71                  \\
                                & EdgePoint2-M32                                       & 63.08                  & 93.46                  & \textcolor{cr}{\textbf{98.85}}         & 37.14                  & \textcolor{cb}{\textbf{71.79}}         & \textcolor{cg}{\textbf{82.14}}         \\
                                & EdgePoint2-L32                                       & \textcolor{cg}{\textbf{66.54}}         & 93.46                  & 97.31                  & \textcolor{cr}{\textbf{40.71}}         & \textcolor{cr}{\textbf{73.93}}         & \textcolor{cg}{\textbf{82.14}}         \\
                                & EdgePoint2-E32                                       & \textcolor{cr}{\textbf{67.31}}         & \textcolor{cr}{\textbf{94.23}}         & \textcolor{cg}{\textbf{98.46}}         & \textcolor{cg}{\textbf{39.64}}         & \textcolor{cg}{\textbf{72.86}}         & \textcolor{cr}{\textbf{82.86}}         \\ \hline
\end{tabular}%
\label{tab:results-hpatches}
\end{table*}

\PAR{Results}
Table \ref{tab:results-hpatches} presents our results categorized by descriptor dimension. The EdgePoint2 series achieve nearly the same accuracy across different descriptor dimensions, even when compared to the high-dimensional descriptors in the illumination change subset. Notably, even the 32-dimensional EdgePoint2 models outperform SuperPoint, the ALIKE series, the ALIKED series, and AWDesc-[T32/CA]. In viewpoint change scenarios, the 64-dimensional EdgePoint2 models remain competitive with the 128-dimensional models. The 48-dimensional EdgePoint2 series perform better than the ALIKE series, and even the 32-dimensional EdgePoint2 series achieve performance comparable to several models with 96 dimension or higher. Considering the number of parameters, the EdgePoint2 series achieves SOTA performance at the lowest cost.

\subsection{Relative Pose Estimation}

\PAR{Setup}
Keypoint detection and description are crucial for relative pose estimation. We utilize MegaDepth-1500~\citep{MegaDepth} and ScanNet-1500~\citep{ScanNet} datasets, which encompass significant illumination and viewpoint changes in both indoor and outdoor environments. Although there are some overlaps between images in MegaDepth-1500 and the training data, our training methodology employs MegaDepth for homography training without relying on image correspondences. Thus, we believe MegaDepth-1500 is suitable for this evaluation. The images are resized to a maximum dimension of 1200 pixels for MegaDepth-1500 and 480 $\times$ 640 for ScanNet-1500. We report the area under the curve (AUC) at thresholds of {5$^\circ$, 10$^\circ$, 20$^\circ$} to measure task accuracy. Additionally, we document the number of inlier points (Inl.N) and the Mean Inlier Ratio (MIR), reflecting the ratio of matching points after RANSAC. We follow the procedure provided in~\citep{LoFTR,EfficientLoFTR}. The optimal threshold for each method is determined as described in~\citep{XFeat}, and LO-RANSAC~\citep{LO-RANSAC} is utilized to estimate the essential matrix. 

We selected the Image Matching Challenge 2022 (IMC2022)~\citep{IMC2022} for further evaluation, as it provides outdoor image pairs for estimating the fundamental matrix. Pose accuracy is assessed using ten thresholds for rotation and translation errors, with metrics that include mean average accuracy for both private and public subsets.

\begin{table*}
\centering
\caption{\textbf{Relative pose estimation on MegaDepth-1500, ScanNet-1500 and IMC2022.} The top three best results of each sector are marked as \textcolor{cr}{red}, \textcolor{cg}{green} and \textcolor{cb}{blue}.}
\resizebox{\textwidth}{!}{%
\begin{tabular}{clcccccccccccc}
\hline
\multirow{2}{*}{\textbf{}}                                   & \multicolumn{1}{c}{\multirow{2}{*}{\textbf{Method}}} & \multicolumn{5}{c}{\textbf{MegaDepth-1500}}                                                                        & \multicolumn{5}{c}{\textbf{ScanNet-1500}}                                                                        & \multicolumn{2}{c}{\textbf{IMC2022}}                            \\
                                                             & \multicolumn{1}{c}{}                                 & AUC@5                          & AUC@10                         & AUC@20                         & Inl.N  & MIR    & AUC@5                          & AUC@10                         & AUC@20                         & Inl.N & MIR   & Private                        & Public                         \\ \hline
\multirow{9}{*}{\rotatebox{90}{\textbf{96 / 128 / 256 dim}}} & SuperPoint~\citep{SuperPoint}                          & 43.52                          & 56.51                          & 66.78                          & 321.1  & 44.34  & 14.67                          & \textcolor{cb}{\textbf{28.97}} & \textcolor{cb}{\textbf{43.76}} & 73.5  & 52.66 & 0.475                          & 0.485                          \\
                                                             & DISK~\citep{DISK}                                      & 54.68                          & 66.63                          & 75.42                          & 1310.5 & 73.44  & 13.61                          & 26.31                          & 39.75                          & 236.2 & 43.92 & \textcolor{cb}{\textbf{0.601}} & 0.589                          \\
                                                             & ALIKE-S~\citep{ALIKE}                                  & 49.12                          & 61.64                          & 71.49                          & 568.9  & 49.32  & 10.72                          & 21.71                          & 33.88                          & 237.4 & 32.04 & 0.471                          & 0.496                          \\
                                                             & ALIKE-N~\citep{ALIKE}                                  & 53.34                          & 65.01                          & 73.95                          & 708.0  & 56.61  & 13.58                          & 26.08                          & 39.51                          & 288.5 & 34.89 & 0.554                          & 0.569                          \\
                                                             & ALIKE-L~\citep{ALIKE}                                  & \textcolor{cb}{\textbf{54.91}} & \textcolor{cb}{\textbf{67.15}} & \textcolor{cb}{\textbf{76.16}} & 772.1  & 56.34  & \textcolor{cb}{\textbf{14.93}} & 28.66                          & 42.47                          & 294.0 & 33.65 & 0.548                          & 0.566                          \\
                                                             & ALIKED-N16~\citep{ALIKED}                              & \textcolor{cg}{\textbf{57.62}} & \textcolor{cg}{\textbf{70.00}} & \textcolor{cg}{\textbf{79.05}} & 743.6  & 67.13  & 13.16                          & 25.90                          & 39.90                          & 87.2  & 49.38 & \textcolor{cr}{\textbf{0.630}} & \textcolor{cr}{\textbf{0.632}} \\
                                                             & ALIKED-N32~\citep{ALIKED}                              & \textcolor{cr}{\textbf{61.10}} & \textcolor{cr}{\textbf{73.56}} & \textcolor{cr}{\textbf{82.16}} & 570.6  & 72.3.0 & 12.72                          & 24.65                          & 37.05                          & 46.6  & 52.05 & \textcolor{cg}{\textbf{0.626}} & \textcolor{cg}{\textbf{0.624}} \\
                                                             & AWDesc-T32~\citep{AWDesc}                              & 51.26                          & 63.84                          & 72.76                          & 298.6  & 50.92  & \textcolor{cr}{\textbf{16.79}} & \textcolor{cr}{\textbf{32.12}} & \textcolor{cr}{\textbf{47.29}} & 64.9  & 56.32 & 0.583                          & \textcolor{cb}{\textbf{0.597}} \\
                                                             & AWDesc-CA~\citep{AWDesc}                               & 53.89                          & 66.66                          & 75.87                          & 302.2  & 55.19  & \textcolor{cg}{\textbf{16.61}} & \textcolor{cg}{\textbf{32.05}} & \textcolor{cg}{\textbf{47.00}} & 62.3  & 57.87 & 0.581                          & 0.595                          \\ \hdashline
\multirow{8}{*}{\rotatebox{90}{\textbf{64 dim}}}             & ALIKE-T~\citep{ALIKE}                                  & 47.23                          & 59.90                          & 69.63                          & 557.7  & 46.40  & 11.22                          & 22.55                          & 35.17                          & 234.7 & 30.62 & 0.487                          & 0.512                          \\
                                                             & ALIKED-T16~\citep{ALIKED}                              & \textcolor{cr}{\textbf{58.46}} & \textcolor{cr}{\textbf{70.89}} & \textcolor{cr}{\textbf{79.77}} & 627.8  & 58.27  & 14.10                          & 27.18                          & 40.54                          & 92.0  & 38.44 & 0.542                          & 0.559                          \\
                                                             & AWDesc-T16~\citep{AWDesc}                              & 48.90                          & 61.27                          & 70.41                          & 255.1  & 48.36  & 14.72                          & 29.41                          & 43.55                          & 53.1  & 54.47 & 0.553                          & 0.574                          \\
                                                             & XFeat~\citep{XFeat}                                    & 43.61                          & 56.78                          & 67.38                          & 852.5  & 52.05  & 12.14                          & 25.08                          & 39.72                          & 411.6 & 40.24 & 0.489                          & 0.505                          \\
                                                             & EdgePoint2-S64                                       & 50.62                          & 62.91                          & 72.06                          & 800.5  & 56.30  & 14.24                          & 27.89                          & 42.83                          & 174.4 & 44.44 & 0.580                          & 0.591                          \\
                                                             & EdgePoint2-M64                                       & 53.20                          & 65.17                          & 74.05                          & 892.4  & 60.04  & \textcolor{cb}{\textbf{15.28}} & \textcolor{cb}{\textbf{29.95}} & \textcolor{cb}{\textbf{44.89}} & 191.6 & 48.66 & \textcolor{cb}{\textbf{0.608}} & \textcolor{cb}{\textbf{0.615}} \\
                                                             & EdgePoint2-L64                                       & \textcolor{cb}{\textbf{54.10}} & \textcolor{cb}{\textbf{66.12}} & \textcolor{cb}{\textbf{74.83}} & 927.3  & 61.42  & \textcolor{cr}{\textbf{16.35}} & \textcolor{cr}{\textbf{31.37}} & \textcolor{cg}{\textbf{46.47}} & 185.2 & 49.69 & \textcolor{cg}{\textbf{0.617}} & \textcolor{cg}{\textbf{0.626}} \\
                                                             & EdgePoint2-E64                                       & \textcolor{cg}{\textbf{54.32}} & \textcolor{cg}{\textbf{66.29}} & \textcolor{cg}{\textbf{75.38}} & 871.8  & 57.83  & \textcolor{cg}{\textbf{16.20}} & \textcolor{cg}{\textbf{31.32}} & \textcolor{cr}{\textbf{46.63}} & 163.6 & 47.50 & \textcolor{cr}{\textbf{0.625}} & \textcolor{cr}{\textbf{0.632}} \\ \hdashline
\multirow{5}{*}{\rotatebox{90}{\textbf{48 dim}}}             & EdgePoint2-T48                                       & 48.71                          & 60.94                          & 70.67                          & 677.5  & 49.77  & 13.72                          & 27.20                          & 42.01                          & 148.9 & 43.29 & 0.550                          & 0.564                          \\
                                                             & EdgePoint2-S48                                       & 49.61                          & 61.79                          & 71.32                          & 749.1  & 51.82  & 14.21                          & 27.79                          & 42.55                          & 170.7 & 46.90 & 0.574                          & 0.585                          \\
                                                             & EdgePoint2-M48                                       & \textcolor{cb}{\textbf{52.07}} & \textcolor{cb}{\textbf{64.38}} & \textcolor{cb}{\textbf{73.30}} & 816.3  & 55.17  & \textcolor{cg}{\textbf{15.32}} & \textcolor{cb}{\textbf{29.22}} & \textcolor{cb}{\textbf{44.11}} & 166.8 & 48.30 & \textcolor{cg}{\textbf{0.603}} & \textcolor{cb}{\textbf{0.609}} \\
                                                             & EdgePoint2-L48                                       & \textcolor{cg}{\textbf{52.55}} & \textcolor{cg}{\textbf{64.48}} & \textcolor{cg}{\textbf{73.63}} & 824.3  & 55.69  & \textcolor{cb}{\textbf{15.20}} & \textcolor{cg}{\textbf{29.62}} & \textcolor{cg}{\textbf{44.41}} & 182.2 & 47.24 & \textcolor{cb}{\textbf{0.602}} & \textcolor{cg}{\textbf{0.610}} \\
                                                             & EdgePoint2-E48                                       & \textcolor{cr}{\textbf{53.19}} & \textcolor{cr}{\textbf{65.43}} & \textcolor{cr}{\textbf{74.60}} & 843.7  & 55.99  & \textcolor{cr}{\textbf{15.86}} & \textcolor{cr}{\textbf{30.80}} & \textcolor{cr}{\textbf{45.54}} & 185.5 & 48.04 & \textcolor{cr}{\textbf{0.614}} & \textcolor{cr}{\textbf{0.614}} \\ \hdashline
\multirow{6}{*}{\rotatebox{90}{\textbf{32 dim}}}             & EdgePoint~\citep{EdgePoint}                            & 42.92                          & 55.72                          & 66.04                          & 264.8  & 50.03  & 11.31                          & 23.46                          & 36.59                          & 52.0  & 59.37 & 0.436                          & 0.460                          \\
                                                             & EdgePoint2-T32                                       & 46.12                          & 58.29                          & 68.11                          & 651.9  & 46.05  & 13.10                          & 25.97                          & 39.76                          & 176.9 & 44.13 & 0.515                          & 0.532                          \\
                                                             & EdgePoint2-S32                                       & 46.24                          & 58.58                          & 68.16                          & 651.8  & 46.50  & 13.46                          & 26.52                          & 40.39                          & 166.6 & 42.76 & 0.527                          & 0.536                          \\
                                                             & EdgePoint2-M32                                       & \textcolor{cb}{\textbf{49.75}} & \textcolor{cb}{\textbf{62.12}} & \textcolor{cb}{\textbf{71.71}} & 748.0  & 50.63  & \textcolor{cb}{\textbf{14.12}} & \textcolor{cb}{\textbf{28.00}} & \textcolor{cg}{\textbf{42.87}} & 157.1 & 40.81 & \textcolor{cb}{\textbf{0.548}} & \textcolor{cb}{\textbf{0.556}} \\
                                                             & EdgePoint2-L32                                       & \textcolor{cr}{\textbf{51.76}} & \textcolor{cr}{\textbf{63.91}} & \textcolor{cg}{\textbf{73.22}} & 744.0  & 51.83  & \textcolor{cg}{\textbf{14.57}} & \textcolor{cg}{\textbf{28.30}} & \textcolor{cb}{\textbf{42.83}} & 158.4 & 45.46 & \textcolor{cg}{\textbf{0.573}} & \textcolor{cg}{\textbf{0.579}} \\
                                                             & EdgePoint2-E32                                       & \textcolor{cr}{\textbf{51.76}} & \textcolor{cg}{\textbf{63.89}} & \textcolor{cr}{\textbf{73.32}} & 778.9  & 52.57  & \textcolor{cr}{\textbf{14.95}} & \textcolor{cr}{\textbf{29.54}} & \textcolor{cr}{\textbf{44.43}} & 159.5 & 45.52 & \textcolor{cr}{\textbf{0.576}} & \textcolor{cr}{\textbf{0.584}} \\ \hline
\end{tabular}%
}
\label{tab:relative-pose-estimation}
\end{table*}

\begin{figure*}[htbp]
    \centering
    \includegraphics[width=\textwidth]{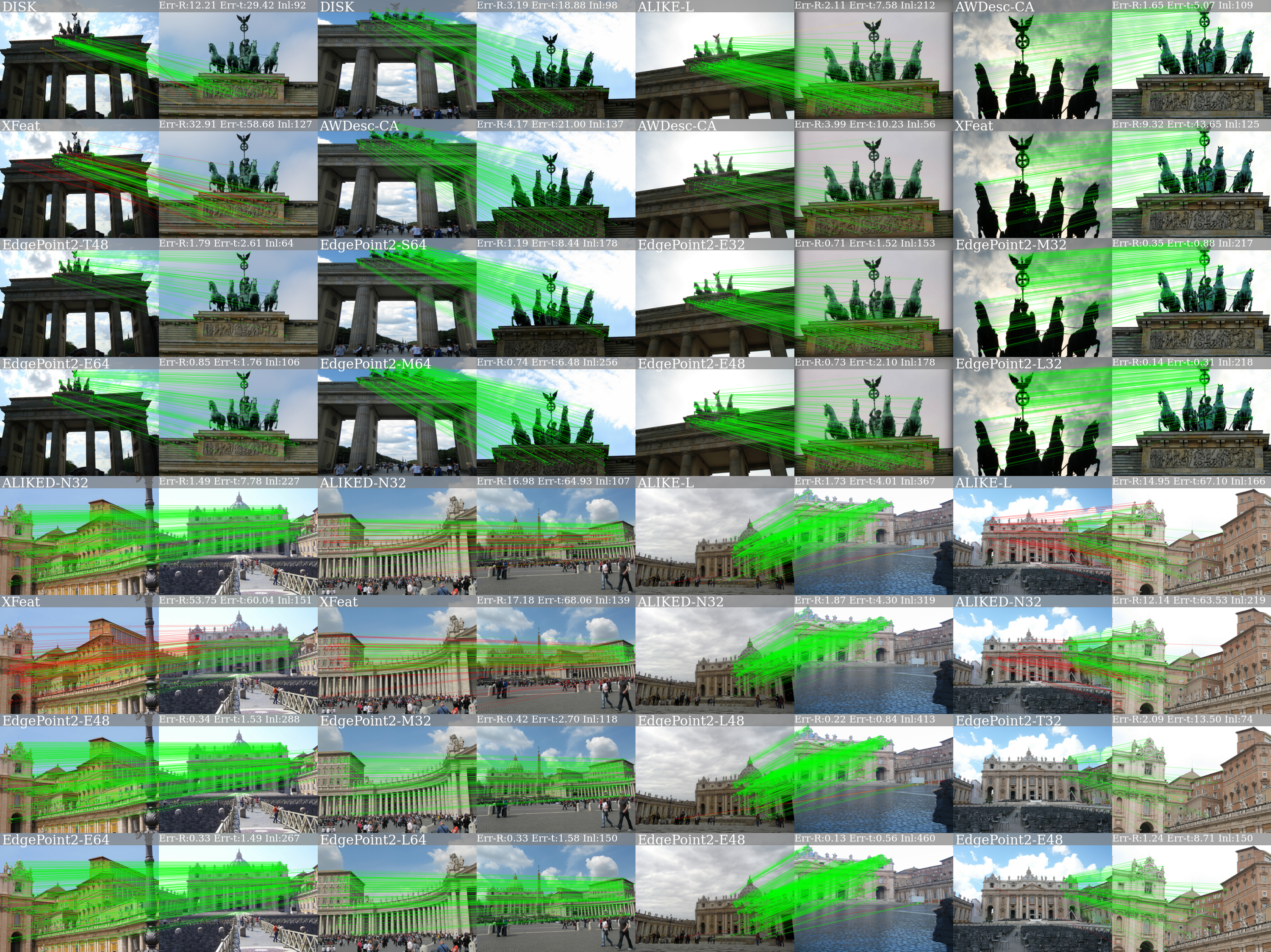}
    \caption{\raggedright \textbf{Qualitative results on MegaDepth-1500.} We choose DISK, XFeat, and the largest models of ALIKE, ALIKED, and AWDesc for comparison. For EdgePoint2, we visualize the results of models of all model sizes (T/S/M/L/E) and dimensions (32/48/64) to demonstrate its consistent performance.}
    \label{fig:visualize-megadepth1500}
\end{figure*}

\begin{figure*}[htbp]
    \centering
    \includegraphics[trim=120 0 140 30, clip, width=\textwidth]{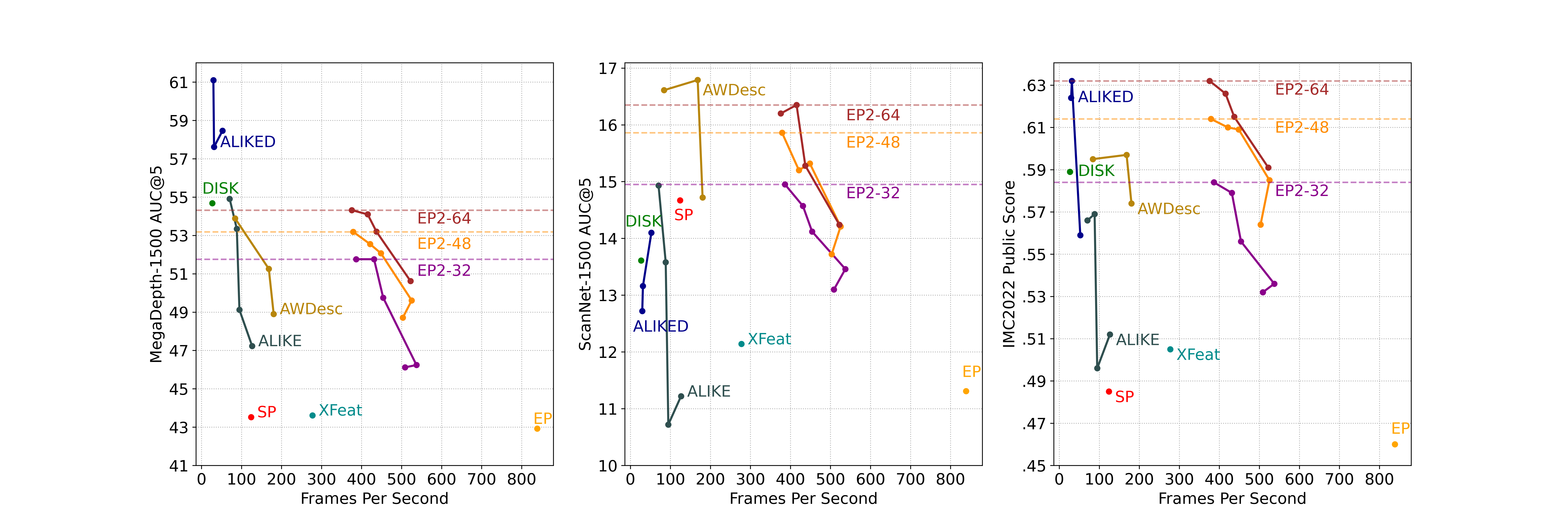}
    \caption{\raggedright \textbf{Comparative visualization of relative pose estimation errors.} Evaluation metrics: AUC@5° (MegaDepth-1500 / ScanNet-1500) and public score (IMC2022).}
    \label{fig:fps_over_relative_pose}
\end{figure*}

\PAR{Results}
The results for relative pose estimation are presented in Table \ref{tab:relative-pose-estimation}. For MegaDepth-1500 dataset, the smallest model, EdgePoint2-T32, surpasses SuperPoint, XFeat, and EdgePoint, while EdgePoint2-M32 outperforms the ALIKE-[T/S] and AWDesc-T16 models with 64- and 96-dimensional descriptors. The EdgePoint2-E64 models demonstrate competitive performance with SOTA methods in terms of the AUC metrics, with performance only slightly behind that of the ALIKED series. Fig.~\ref{fig:visualize-megadepth1500} shows the superior performance of EdgePoint2 compared to other methods. n the ScanNet-1500 dataset, even the EdgePoint2-E32 model outperforms other models, including the ALIKED series, while utilizing only 32-dimensional descriptors. The accuracy of the EdgePoint2-E64 model is just slightly lower than that of AWDesc-[T32/CA], but it operates at least twice as fast. In the final IMC2022 evaluation, EdgePoint2-E64 reaches SOTA alongside ALIKED[-N16/N32], outperforming all other models. The EdgePoint2 32-dimensional models outperform XFeat, EdgePoint, and ALIKE-[T/S] in both private and public scores, while the 48-dimensional models also exceed the performance of DISK, ALIKE series and AWDesc series. Throughout the relative pose estimation experiments, the EdgePoint2 series models exhibit superior performance while utilizing fewer parameters and lower descriptor dimensions, as shown in Fig. \ref{fig:fps_over_relative_pose}.

\subsection{FM-Bench}

\begin{table*}[]
\centering
\caption{\textbf{FM-Bench result on TUM, KITTI, T\&T and CPC.} The top three best results of each sector are marked as \textcolor{cr}{red}, \textcolor{cg}{green} and \textcolor{cb}{blue}.}
\resizebox{\textwidth}{!}{%
\begin{tabular}{clcccccccccccc}
\hline
\multirow{2}{*}{}               & \multicolumn{1}{c}{\multirow{2}{*}{\textbf{Method}}} & \multicolumn{3}{c}{\textbf{TUM}}  & \multicolumn{3}{c}{\textbf{KITTI}} & \multicolumn{3}{c}{\textbf{T\&T}} & \multicolumn{3}{c}{\textbf{CPC}}  \\ 
                                & \multicolumn{1}{c}{}                                 & Recall & Inl.R(-m)    & Corrs(-m) & Recall  & Inl.R(-m)    & Corrs(-m) & Recall & Inl.R(-m)    & Corrs(-m) & Recall & Inl.R(-m)    & Corrs(-m) \\ \hline
\multirow{9}{*}{\rotatebox{90}{\textbf{96 / 128 / 256 dim}}} & SuperPoint\cite{SuperPoint}          & 44.00  & 72.24(62.23) & 43(249)   & 85.10   & 98.06(91.28) & 74(411)   & 84.00  & 84.46(71.16) & 57(615)   & 42.50  & 77.71(67.08) & 38(283)   \\
                                & DISK\cite{DISK}                                      & \textcolor{cg}{\textbf{59.20}}  & 74.36(67.90) & 236(1299) & 89.40   & 98.71(97.38) & 422(2042) & 78.20  & 83.53(79.75) & 90(608)   & \textcolor{cb}{\textbf{50.40}}  & 90.67(87.42) & 100(619)  \\
                                & ALIKE-S\cite{ALIKE}                                  & 51.90  & 72.11(62.10) & 152(754)  & 89.90   & 98.81(94.96) & 241(1022) & 77.30  & 80.84(73.99) & 51(350)   & 36.20  & 84.94(80.63) & 41(161)   \\
                                & ALIKE-N\cite{ALIKE}                                  & 52.70  & 73.39(64.90) & 194(945)  & 89.40   & 98.69(96.13) & 250(1056) & 83.00  & 83.79(78.06) & 57(382)   & 36.00  & 88.66(85.84) & 52(212)   \\
                                & ALIKE-L\cite{ALIKE}                                  & 53.10  & 72.85(63.34) & 195(990)  & \textcolor{cb}{\textbf{90.70}}   & 98.66(96.07) & 259(1132) & 85.40  & 85.08(78.57) & 62(414)   & 41.60  & 89.31(86.04) & 54(233)   \\
                                & ALIKED-N16\cite{ALIKED}                              & \textcolor{cr}{\textbf{61.90}}  & 75.47(69.84) & 81(314)   & \textcolor{cg}{\textbf{92.30}}   & 98.56(95.98) & 276(795)  & \textcolor{cr}{\textbf{91.80}}  & 87.08(81.11) & 137(731)  & \textcolor{cr}{\textbf{60.20}}  & 90.94(85.72) & 114(558)  \\
                                & ALIKED-N32\cite{ALIKED}                              & \textcolor{cb}{\textbf{55.20}}  & 75.53(70.24) & 59(215)   & \textcolor{cr}{\textbf{92.40}}   & 98.84(96.02) & 170(469)  & \textcolor{cg}{\textbf{91.40}}  & 87.22(81.32) & 127(707)  & \textcolor{cg}{\textbf{59.10}}  & 91.72(86.80) & 112(557)  \\
                                & AWDesc-T32\cite{AWDesc}                              & 45.80  & 74.04(68.78) & 46(220)   & 88.40   & 98.45(95.47) & 88(410)   & \textcolor{cb}{\textbf{87.10}}  & 85.68(79.96) & 65(540)   & 47.80  & 87.53(84.07) & 44(261)   \\
                                & AWDesc-CA\cite{AWDesc}                               & 45.70  & 73.27(67.64) & 44(207)   & 87.10   & 98.48(94.86) & 79(364)   & 84.20  & 86.45(81.17) & 58(469)   & 44.90  & 88.36(85.08) & 40(242)   \\ \hdashline
\multirow{8}{*}{\rotatebox{90}{\textbf{64 dim}}} & ALIKE-T\cite{ALIKE}                          & 51.60  & 72.16(60.62) & 185(991)  & \textcolor{cg}{\textbf{89.20}}   & 98.65(93.56) & 219(999)  & 78.40  & 82.18(71.53) & 48(361)   & 38.20  & 83.99(77.04) & 41(172)   \\
                                & ALIKED-T16\cite{ALIKED}                              & 55.40  & 75.31(68.41) & 88(422)   & \textcolor{cr}{\textbf{92.50}}   & 98.50(95.78) & 353(1109) & \textcolor{cr}{\textbf{91.60}}  & 86.10(79.58) & 119(685)  & \textcolor{cr}{\textbf{57.50}}  & 89.49(83.80) & 93(453)   \\
                                & AWDesc-T16\cite{AWDesc}                              & 46.10  & 74.00(69.15) & 46(209)   & 88.20   & 98.51(95.51) & 86(395)   & 87.40  & 84.73(78.84) & 62(504)   & 46.50  & 86.80(82.86) & 41(241)   \\
                                & XFeat\cite{XFeat}                                    & 41.80  & 70.42(60.63) & 153(935)  & 80.30   & 98.56(92.77) & 180(1068) & 67.50  & 82.26(72.83) & 35(406)   & 28.10  & 81.50(72.85) & 36(249)   \\
                                & EdgePoint2-S64                                       & \textcolor{cg}{\textbf{56.80}}  & 75.00(67.83) & 165(772)  & 87.80   & 98.78(96.08) & 342(1509) & 86.70  & 84.91(76.63) & 85(620)   & \textcolor{cb}{\textbf{47.20}}  & 86.00(80.04) & 57(306)   \\
                                & EdgePoint2-M64                                       & \textcolor{cb}{\textbf{55.60}}  & 75.38(68.09) & 168(773)  & \textcolor{cb}{\textbf{88.70}}   & 98.76(96.13) & 340(1514) & 87.20  & 84.46(76.09) & 83(607)   & 45.20  & 85.49(79.28) & 55(286)   \\
                                & EdgePoint2-L64                                       & 54.90  & 75.22(67.92) & 170(790)  & 88.60   & 98.77(95.89) & 349(1546) & \textcolor{cg}{\textbf{88.60}}  & 84.57(75.29) & 87(643)   & \textcolor{cb}{\textbf{47.20}}  & 85.36(77.97) & 58(316)   \\
                                & EdgePoint2-E64                                       & \textcolor{cr}{\textbf{56.90}}  & 75.00(67.94) & 163(743)  & 87.50   & 98.78(96.11) & 331(1453) & \textcolor{cg}{\textbf{88.60}}  & 84.78(76.67) & 88(642)   & \textcolor{cg}{\textbf{48.60}}  & 85.89(79.83) & 59(314)   \\ \hdashline
\multirow{5}{*}{\rotatebox{90}{\textbf{48 dim}}} & EdgePoint2-T48                               & \textcolor{cb}{\textbf{55.70}}  & 74.93(67.15) & 163(764)  & \textcolor{cg}{\textbf{87.80}}   & 98.85(95.06) & 335(1491) & 85.30  & 83.79(70.24) & 81(629)   & 44.70  & 78.83(67.23) & 50(284)   \\
                                & EdgePoint2-S48                                       & \textcolor{cr}{\textbf{55.80}}  & 75.22(67.87) & 164(748)  & 87.40   & 98.67(95.90) & 333(1479) & \textcolor{cg}{\textbf{87.70}}  & 84.84(75.32) & 84(627)   & \textcolor{cr}{\textbf{47.60}}  & 84.44(77.34) & 55(297)   \\
                                & EdgePoint2-M48                                       & 55.40  & 75.19(68.00) & 163(755)  & \textcolor{cr}{\textbf{88.00}}   & 98.85(96.15) & 337(1479) & \textcolor{cb}{\textbf{86.10}}  & 84.76(76.64) & 85(615)   & 45.40  & 85.98(79.58) & 57(298)   \\
                                & EdgePoint2-L48                                       & \textcolor{cg}{\textbf{57.10}}  & 75.20(68.18) & 165(762)  & \textcolor{cb}{\textbf{87.60}}   & 98.79(95.86) & 344(1524) & 85.90  & 84.25(74.99) & 82(616)   & \textcolor{cb}{\textbf{46.30}}  & 84.61(77.20) & 55(299)   \\
                                & EdgePoint2-E48                                       & \textcolor{cb}{\textbf{55.70}}  & 75.88(68.00) & 164(760)  & 87.40   & 98.71(95.87) & 340(1504) & \textcolor{cr}{\textbf{88.00}}  & 84.64(75.31) & 85(631)   & \textcolor{cg}{\textbf{47.10}}  & 84.79(78.04) & 56(304)   \\ \hdashline
\multirow{6}{*}{\rotatebox{90}{\textbf{32 dim}}} & EdgePoint\cite{EdgePoint}                    & 47.50  & 73.70(64.82) & 53(291)   & \textcolor{cg}{\textbf{87.60}}   & 98.14(90.70) & 108(625)  & 78.80  & 80.28(63.11) & 58(641)   & 34.00  & 67.17(53.46) & 34(261)   \\
                                & EdgePoint2-T32                                       & 54.20  & 75.22(67.00) & 166(797)  & 87.00   & 98.83(94.82) & 333(1483) & 83.00  & 82.55(68.87) & 76(580)   & 38.50  & 74.27(62.67) & 44(237)   \\
                                & EdgePoint2-S32                                       & \textcolor{cr}{\textbf{56.90}}  & 74.85(67.49) & 162(740)  & 86.40   & 98.81(95.80) & 318(1380) & 84.40  & 83.07(73.49) & 77(561)   & 43.00  & 80.22(73.07) & 47(239)   \\
                                & EdgePoint2-M32                                       & \textcolor{cg}{\textbf{55.50}}  & 75.16(67.20) & 162(771)  & \textcolor{cr}{\textbf{88.20}}   & 98.73(94.99) & 341(1527) & \textcolor{cg}{\textbf{85.50}}  & 83.26(69.89) & 78(604)   & \textcolor{cr}{\textbf{44.20}}  & 79.39(67.72) & 50(284)   \\
                                & EdgePoint2-L32                                       & \textcolor{cb}{\textbf{54.90}}  & 75.13(67.60) & 165(765)  & \textcolor{cb}{\textbf{87.20}}   & 98.84(95.84) & 329(1439) & \textcolor{cr}{\textbf{85.80}}  & 84.28(74.53) & 79(586)   & \textcolor{cg}{\textbf{44.10}}  & 81.74(74.94) & 50(258)   \\
                                & EdgePoint2-E32                                       & 53.10  & 75.04(67.65) & 172(800)  & 86.70   & 98.86(96.09) & 335(1466) & \textcolor{cb}{\textbf{85.10}}  & 84.14(75.71) & 78(566)   & \textcolor{cb}{\textbf{43.70}}  & 83.63(77.19) & 50(251)   \\ \hline
\end{tabular}%
}
\label{tab:fmbench}
\end{table*}

\PAR{Setup} 
FM-Bench~\citep{FM-Bench} serves as a benchmark for evaluating keypoint detection and descriptor performance. We assess four datasets representing different scenarios: the TUM dataset~\citep{TUM} for indoor SLAM, the KITTI dataset~\citep{KITTI} for autonomous driving, the Tanks and Templates dataset (T\&T)~\citep{TT} for wide-baseline reconstruction, and the CPC~\citep{CPC} for wild reconstruction from web images. Each dataset consists of 1000 images and their corresponding ground-truth fundamental matrices.

Following the evaluation procedure outlined in~\citep{ASLFeat}, we assess three metrics: recall, inliers ratio (Inl.R(-m)), and the number of feature correspondences (Corrs(-m)). Recall indicates the percentage of pose estimations with distance errors under a default threshold of 0.05. The inlier ratio reflects the ratio of inliers to matches, while Corrs represents the count of matches, with the suffix -m indicating the count before RANSAC outlier rejection.

\PAR{Result} 
Results are summarized in Table \ref{tab:fmbench}. The EdgePoint2 series exhibits recall rates slightly lower than those of the ALIKED series across most datasets, while outperforming other SOTA methods such as DISK, ALIKE, and AWDesc. Among lightweight models, EdgePoint2 demonstrates advantages over both XFeat and EdgePoint. On the TUM and KITTI datasets, all EdgePoint2 models show competitive performance, indicating that the 32-dim models are excellent choices for applications involving continuous image streams.

\subsection{Visual Localization}

\begin{table*}
\centering
\caption{\textbf{Visual localization on Aachen Day-Night v1.1 and InLoc.} The top three best results of each sector are marked as \textcolor{cr}{red}, \textcolor{cg}{green} and \textcolor{cb}{blue}.}
\resizebox{0.9\textwidth}{!}{%
\begin{tabular}{clcccccccccccc}
\hline
\multirow{4}{*}{}               & \multicolumn{1}{c}{\multirow{4}{*}{\textbf{Method}}} & \multicolumn{6}{c}{{\ul \textbf{Aachen Day-Night v1.1}}}                & \multicolumn{6}{c}{{\ul \textbf{InLoc}}}                                \\
                                & \multicolumn{1}{c}{}                                 & \multicolumn{3}{c}{Day}            & \multicolumn{3}{c}{Night}          & \multicolumn{3}{c}{DUC1}           & \multicolumn{3}{c}{DUC2}           \\ 
                                & \multicolumn{1}{c}{}                                 & 0.25$m$   & 0.5$m$    & 5$m$       & 0.25$m$   & 0.5$m$    & 5$m$       & 0.25$m$   & 0.5$m$    & 5$m$       & 0.25$m$   & 0.5$m$    & 5$m$       \\
                                & \multicolumn{1}{c}{}                                 & 2$^\circ$ & 5$^\circ$ & 10$^\circ$ & 2$^\circ$ & 5$^\circ$ & 10$^\circ$ & 2$^\circ$ & 5$^\circ$ & 10$^\circ$ & 2$^\circ$ & 5$^\circ$ & 10$^\circ$ \\ \hline
\multirow{9}{*}{\rotatebox{90}{\textbf{96 / 128 / 256 dim}}} & SuperPoint~\citep{SuperPoint}          & \textcolor{cb}{\textbf{88.3}}      & 94.4      & 98.1       & 69.1      & 85.9      & 95.8       & 33.3      & 50.0      & 59.6       & 32.8      & \textcolor{cb}{\textbf{51.9}}      & 63.4       \\
                                & DISK~\citep{DISK}                                      & 87.3      & \textcolor{cb}{\textbf{95.5}}      & \textcolor{cg}{\textbf{98.5}}       & \textcolor{cr}{\textbf{78.0}}      & 89.0      & \textcolor{cb}{\textbf{99.0}}       & 35.9      & 53.0      & 66.2       & 24.4      & 40.5      & 57.3       \\
                                & ALIKE-S~\citep{ALIKE}                                  & 87.3      & 93.7      & 97.8       & 69.6      & 86.9      & 95.8       & 32.3      & 49.5      & 59.6       & 23.7      & 38.2      & 46.4       \\
                                & ALIKE-N~\citep{ALIKE}                                  & 88.2      & 94.3      & 97.9       & 71.7      & 88.5      & \textcolor{cb}{\textbf{99.0}}       & 32.3      & 48.5      & 60.6       & 32.1      & 41.2      & 51.1       \\
                                & ALIKE-L~\citep{ALIKE}                                  & \textcolor{cb}{\textbf{88.3}}      & 95.1      & \textcolor{cg}{\textbf{98.5}}       & 72.8      & 89.0      & \textcolor{cr}{\textbf{99.5}}       & 32.3      & 48.5      & 58.1       & 28.2      & 42.0      & 46.6       \\
                                & ALIKED-N16~\citep{ALIKED}                              & \textcolor{cr}{\textbf{88.8}}      & \textcolor{cr}{\textbf{96.1}}      & \textcolor{cr}{\textbf{98.8}}       & 74.3      & 89.5      & \textcolor{cb}{\textbf{99.0}}       & 32.8      & \textcolor{cb}{\textbf{54.0}}      & 65.7       & \textcolor{cb}{\textbf{35.1}}      & 50.4      & 58.0       \\
                                & ALIKED-N32~\citep{ALIKED}                              & 86.4      & 95.3      & 98.3       & \textcolor{cb}{\textbf{74.9}}      & \textcolor{cg}{\textbf{90.1}}      & \textcolor{cr}{\textbf{99.5}}       & \textcolor{cb}{\textbf{36.9}}      & 51.0      & \textcolor{cb}{\textbf{66.7}}       & \textcolor{cg}{\textbf{35.9}}      & 49.6      & \textcolor{cb}{\textbf{64.1}}       \\
                                & AWDesc-T32~\citep{AWDesc}                              & 88.2      & \textcolor{cg}{\textbf{95.9}}      & 98.4       & \textcolor{cg}{\textbf{77.0}}      & \textcolor{cg}{\textbf{90.1}}      & 98.4       & \textcolor{cg}{\textbf{38.4}}      & \textcolor{cg}{\textbf{57.6}}      & \textcolor{cg}{\textbf{68.7}}       & 34.4      & \textcolor{cg}{\textbf{55.7}}      & \textcolor{cg}{\textbf{65.6}}       \\
                                & AWDesc-CA~\citep{AWDesc}                               & \textcolor{cg}{\textbf{88.5}}      & 95.4      & 98.4       & \textcolor{cb}{\textbf{74.9}}      & \textcolor{cr}{\textbf{91.1}}      & \textcolor{cb}{\textbf{99.0}}       & \textcolor{cr}{\textbf{39.9}}      & \textcolor{cr}{\textbf{60.1}}      & \textcolor{cr}{\textbf{71.2}}       & \textcolor{cr}{\textbf{42.7}}      & \textcolor{cr}{\textbf{58.0}}      & \textcolor{cr}{\textbf{68.7}}       \\ \hdashline
\multirow{8}{*}{\rotatebox{90}{\textbf{64 dim}}} & ALIKE-T~\citep{ALIKE}                          & 86.4      & 94.3      & 97.9       & 69.1      & 86.4      & 95.8       & 31.3      & 45.5      & 56.1       & 26.0      & 38.9      & 47.3       \\
                                & ALIKED-T16~\citep{ALIKED}                              & \textcolor{cg}{\textbf{89.1}}      & \textcolor{cg}{\textbf{95.0}}      & 98.2       & \textcolor{cg}{\textbf{77.5}}      & \textcolor{cg}{\textbf{91.1}}      & \textcolor{cr}{\textbf{99.0}}       & 33.8      & 49.0      & 59.1       & 33.6      & 49.6      & 57.3       \\
                                & AWDesc-T16~\citep{AWDesc}                              & \textcolor{cr}{\textbf{89.2}}      & \textcolor{cr}{\textbf{95.1}}      & \textcolor{cr}{\textbf{98.4}}       & 75.4      & 88.0      & 96.9       & \textcolor{cr}{\textbf{38.9}}      & \textcolor{cg}{\textbf{56.1}}      & \textcolor{cb}{\textbf{66.7}}       & 35.1      & 51.9      & 64.9       \\
                                & XFeat~\citep{XFeat}                                    & 86.0      & 93.9      & 97.6       & 70.7      & 85.9      & 97.9       & 36.4      & 54.0      & 64.1       & 30.5      & 44.3      & 57.3       \\
                                & EdgePoint2-S64                                       & 87.1      & 94.7      & 98.1       & 74.3      & 89.5      & 97.9       & 35.4      & 50.0      & 64.1       & \textcolor{cr}{\textbf{42.7}}      & \textcolor{cg}{\textbf{55.0}}      & 65.6       \\
                                & EdgePoint2-M64                                       & 87.1      & 94.7      & 98.2       & 75.9      & \textcolor{cb}{\textbf{90.1}}      & \textcolor{cr}{\textbf{99.0}}       & \textcolor{cb}{\textbf{37.9}}      & \textcolor{cr}{\textbf{57.1}}      & \textcolor{cr}{\textbf{68.7}}       & 39.7      & \textcolor{cr}{\textbf{60.3}}      & \textcolor{cb}{\textbf{66.4}}       \\
                                & EdgePoint2-L64                                       & \textcolor{cb}{\textbf{87.3}}      & \textcolor{cg}{\textbf{95.0}}      & \textcolor{cg}{\textbf{98.3}}       & \textcolor{cr}{\textbf{78.5}}      & 89.0      & \textcolor{cr}{\textbf{99.0}}       & \textcolor{cr}{\textbf{38.9}}      & \textcolor{cg}{\textbf{56.1}}      & \textcolor{cg}{\textbf{68.2}}       & \textcolor{cb}{\textbf{40.5}}      & \textcolor{cg}{\textbf{55.0}}      & \textcolor{cr}{\textbf{67.9}}       \\
                                & EdgePoint2-E64                                       & 86.9      & \textcolor{cg}{\textbf{95.0}}      & \textcolor{cg}{\textbf{98.3}}       & \textcolor{cb}{\textbf{77.0}}      & \textcolor{cr}{\textbf{91.6}}      & \textcolor{cr}{\textbf{99.0}}       & 36.4      & 54.5      & 66.2       & \textcolor{cg}{\textbf{41.2}}      & \textcolor{cb}{\textbf{58.8}}      & \textcolor{cg}{\textbf{67.2}}       \\ \hdashline
\multirow{5}{*}{\rotatebox{90}{\textbf{48 dim}}} & EdgePoint2-T48                                       & 86.8      & 94.1      & \textcolor{cb}{\textbf{98.1}}       & 74.9      & 87.4      & \textcolor{cb}{\textbf{97.9}}       & 34.8      & 48.5      & 61.6       & 37.4      & \textcolor{cb}{\textbf{55.7}}      & 65.6       \\
                                & EdgePoint2-S48                                       & 87.1      & 94.4      & \textcolor{cb}{\textbf{98.1}}       & \textcolor{cg}{\textbf{77.0}}      & 89.5      & \textcolor{cb}{\textbf{97.9}}       & 30.3      & 49.0      & 59.6       & \textcolor{cb}{\textbf{38.9}}      & 54.2      & 63.2       \\
                                & EdgePoint2-M48                                       & \textcolor{cr}{\textbf{88.1}}      & \textcolor{cb}{\textbf{94.9}}      & \textcolor{cb}{\textbf{98.1}}       & \textcolor{cr}{\textbf{78.5}}      & \textcolor{cr}{\textbf{91.6}}      & \textcolor{cg}{\textbf{98.4}}       & \textcolor{cb}{\textbf{35.9}}      & \textcolor{cg}{\textbf{54.5}}      & \textcolor{cb}{\textbf{65.7}}       & 36.6      & 55.0      & \textcolor{cr}{\textbf{67.9}}       \\
                                & EdgePoint2-L48                                       & \textcolor{cg}{\textbf{88.0}}      & \textcolor{cr}{\textbf{95.5}}      & \textcolor{cr}{\textbf{98.3}}       & 75.4      & \textcolor{cb}{\textbf{90.1}}      & \textcolor{cr}{\textbf{99.0}}       & \textcolor{cr}{\textbf{37.9}}      & \textcolor{cr}{\textbf{57.1}}      & \textcolor{cr}{\textbf{68.7}}       & \textcolor{cg}{\textbf{39.7}}      & \textcolor{cr}{\textbf{60.3}}      & \textcolor{cg}{\textbf{66.4}}       \\
                                & EdgePoint2-E48                                       & \textcolor{cg}{\textbf{88.0}}      & \textcolor{cg}{\textbf{95.0}}      & \textcolor{cr}{\textbf{98.3}}       & \textcolor{cb}{\textbf{76.4}}      & \textcolor{cr}{\textbf{91.6}}      & \textcolor{cb}{\textbf{97.9}}       & \textcolor{cr}{\textbf{37.9}}      & \textcolor{cb}{\textbf{53.0}}      & \textcolor{cg}{\textbf{66.2}}       & \textcolor{cr}{\textbf{41.2}}      & \textcolor{cg}{\textbf{58.0}}      & \textcolor{cg}{\textbf{66.4}}       \\ \hdashline
\multirow{6}{*}{\rotatebox{90}{\textbf{32 dim}}} & EdgePoint~\citep{EdgePoint}                    & 86.9      & 93.2      & 97.6       & 66.0      & 85.9      & 95.3       & 28.3      & 43.4      & 53.0       & 22.9      & 37.4      & 49.6       \\
                                & EdgePoint2-T32                                       & 86.8      & 94.3      & \textcolor{cb}{\textbf{97.9}}       & 70.2      & 86.9      & 95.3       & 31.3      & 46.5      & 60.6       & 37.4      & 54.2      & 62.6       \\
                                & EdgePoint2-S32                                       & 85.9      & 93.9      & 97.7       & 70.7      & 85.3      & 96.3       & \textcolor{cb}{\textbf{33.3}}      & \textcolor{cg}{\textbf{50.5}}      & 59.6       & \textcolor{cb}{\textbf{38.9}}      & 55.0      & \textcolor{cb}{\textbf{64.1}}       \\
                                & EdgePoint2-M32                                       & \textcolor{cg}{\textbf{87.0}}      & \textcolor{cg}{\textbf{94.7}}      & \textcolor{cg}{\textbf{98.2}}       & \textcolor{cb}{\textbf{72.3}}      & \textcolor{cr}{\textbf{89.0}}      & \textcolor{cb}{\textbf{97.4}}       & \textcolor{cg}{\textbf{33.8}}      & 49.5      & \textcolor{cb}{\textbf{61.6}}       & \textcolor{cr}{\textbf{41.2}}      & \textcolor{cg}{\textbf{57.3}}      & \textcolor{cr}{\textbf{66.4}}       \\
                                & EdgePoint2-L32                                       & \textcolor{cg}{\textbf{87.0}}      & \textcolor{cb}{\textbf{94.4}}      & \textcolor{cb}{\textbf{97.9}}       & \textcolor{cr}{\textbf{75.4}}      & \textcolor{cr}{\textbf{89.0}}      & \textcolor{cg}{\textbf{97.9}}       & \textcolor{cr}{\textbf{36.4}}      & \textcolor{cr}{\textbf{51.5}}      & \textcolor{cr}{\textbf{67.7}}       & \textcolor{cg}{\textbf{40.5}}      & \textcolor{cg}{\textbf{57.3}}      & \textcolor{cg}{\textbf{64.9}}       \\
                                & EdgePoint2-E32                                       & \textcolor{cr}{\textbf{88.1}}      & \textcolor{cr}{\textbf{94.8}}      & \textcolor{cr}{\textbf{98.3}}       & \textcolor{cr}{\textbf{75.4}}      & \textcolor{cb}{\textbf{88.5}}      & \textcolor{cr}{\textbf{99.0}}       & 32.8      & \textcolor{cb}{\textbf{50.0}}      & \textcolor{cg}{\textbf{67.2}}       & 35.1      & \textcolor{cr}{\textbf{59.5}}      & 62.6       \\ \hline
\end{tabular}%
}
\label{tab:hloc}
\end{table*}

\PAR{Setup} 
We implement the Hierarchical Localization framework outlined in~\citep{HLoc}. For evaluation, we select two datasets: the Aachen Day-Night v1.1~\citep{HLocDataset} to assess the impact of illumination changes between day and night, and the InLoc~\citep{InLoc} dataset for large-scale indoor localization under varying viewpoints and moving objects.

We resize the images to a maximum dimension of 1024 pixels for Aachen and 1600 pixels for InLoc, adhering to the default configurations recommended by HLoc. For evaluation metrics, we use standard HLoc metrics, estimating camera pose accuracy under thresholds of {0.25m, 0.5m, 5m} for distance and {2$^\circ$, 5$^\circ$, 10$^\circ$} for orientation.

\PAR{Results}
The results are presented in Table \ref{tab:hloc}. In the Aachen Day-Night v1.1 dataset, all models exhibit similar performance during daytime conditions. In night time conditions, the EdgePoint2-L64 and EdgePoint2-E64 models achieve SOTA results, while the EdgePoint2 48-dim models also demonstrate strong performance, outperforming most SOTA models. Moreover, even the EdgePoint2 32-dim models achieve better results than both SuperPoint and XFeat. For the InLoc dataset, the EdgePoint2 64-dim model scores slightly lower than the AWDesc-CA, but all EdgePoint2 models perform exceptionally well under DUC2 conditions. Throughout these experiments, EdgePoint2 showcases robust performance across both day-night and complex scenarios, with even the smallest model, EdgePoint2-T32, outperforming SuperPoint and EdgePoint in most situations, thereby highlighting the effectiveness of the EdgePoint2 series across various configurations.

\subsection{Ablation Study}

\begin{table*}
\caption{\raggedright \textbf{Ablation studies of description distillation loss.} The best result of each sector is marked as \textbf{bold}.}
\centering
\begin{tabular}{clccc}
\hline
\multicolumn{2}{l}{\multirow{2}{*}{Distillation}}                                & \multicolumn{3}{c}{AUC}                          \\
\multicolumn{2}{l}{}                                                             & @5             & @10            & @20            \\ \hline
(a) & Similarity-Preserving~\citep{Similarity-Preserving}                          & 44.53          & 56.14          & 65.67          \\
(b) & AWDesc~\citep{AWDesc}                                                        & 51.18          & 62.92          & 72.07          \\
(c) & HF-Net~\citep{HF-Net} w/. PCA                                                & 50.43          & 62.52          & 71.82          \\
(d) & HF-Net~\citep{HF-Net} w/. LRA                                                & 48.13          & 60.04          & 69.18          \\
(e) & $L_{\mathrm{op}}$ w/. PCA, $N$=1, w/o. $L_{\mathrm{sim}}$~\citep{EdgePoint} & 50.89          & 62.74          & 71.80          \\
(f) & $L_{\mathrm{op}}$ w/. LRA, $N$=1, w/o. $L_{\mathrm{sim}}$                  & \textbf{51.75} & \textbf{63.95} & \textbf{72.96} \\ \hline
(g) & $L_{\mathrm{op}}$ w/. LRA, $N$=4, w/o. $L_{\mathrm{sim}}$                  & \textbf{52.17} & 64.15          & 73.06          \\
(h) & $L_{\mathrm{op}}$ w/. LRA, $N$=4, w/. $L_{\mathrm{sim}}$                   & 52.07          & \textbf{64.38} & \textbf{73.30} \\ \hline
\end{tabular}
\label{tab:ablation-study}
\end{table*}

\PAR{Setup} We train our medium-sized EdgePoint2-M48 network with various loss configurations and evaluate on MegaDepth-1500 for ablation studies. We compare the proposed Orthogonal Procrustes loss with previous descriptor distillation methods including Similarity-Preserving~\citep{Similarity-Preserving}, AWDesc~\citep{AWDesc}, HF-Net~\citep{HF-Net} and EdgePoint~\citep{EdgePoint}. Since HF-Net cannot achieve cross dimension distillation, we compress the descriptors based on the entire training data. Moreover, we assess the impact of compression methods, including PCA and LRA, for different loss functions. Additionally, we also assess the effectiveness of enhanced image set and analyze the utility of the similarity loss. 

\PAR{Results} The ablation study results are shown in Table \ref{tab:ablation-study}. In the comparison of experiments (a)-(f), the proposed Orthogonal Procrustes loss with LRA achieved the highest accuracy in all metrics. Experiments (c)-(f) specifically compare the effect of different compression methods under different loss functions. Since HF-Net was trained with compression applied to the entire dataset, the number of descriptors is large, making PCA—which removes overall bias and centralizes data—a more effective method than LRA for distinguishing all descriptors. Conversely, in the Orthogonal Procrustes loss, compression is performed on small batches of descriptors, with the number of descriptors matching the target dimension. In contrast to the descriptor distillation approach introduced in EdgePoint~\citep{EdgePoint} using PCA, selecting LRA for compression enables the lossless preservation of the relative relationships among all descriptors. This leads to differences in compression results using PCA and LRA between HF-Net and our Orthogonal Procrustes loss.

By incorporating image augmentation into our Orthogonal Procrustes loss, as demonstrated in experiment (g), one can observe an improvement in all model metrics. This enhancement underscores the effectiveness of image augmentation in refining model performance. Furthermore, in experiment (h), we introduced $L_\mathrm{sim}$, which represents our final method. This addition leads the model to achieve the highest accuracy in average, highlighting the benefit of enforcing absolute distribution consistency. These results collectively illustrate the significant impact of combining image augmentation and absolute distribution consistency.

\section{Conclusions}

In this work, we present EdgePoint2, a series of lightweight models designed for edge computing applications. By introducing enhanced Orthogonal Procrustes loss and similarity loss during training, EdgePoint2 effectively generates the most compact descriptors for now. The models demonstrate significant advantages in flexibility and versatility, enabled by its various network configurations (ranging from Tiny to Enormous) and descriptor dimensionalities (32/48/64). Across all experimental results, our method has achieved SOTA performance on 10 datasets spanning 4 different tasks, demonstrating the consistent and robust performance of our approach across a diverse range of challenges. Notably, while attaining this level of accuracy, our method utilizes the minimal number of parameters and computational resources. Our method does not rely on any specialized network architectures, which not only simplifies the model but also enables ultra-high inference speeds.

Although EdgePoint2 demonstrates outstanding performance, it still exhibits limitations in handling image matching under extreme viewpoint and scale variations. With attention-based matchers showing great improvements in sparse keypoint matching, we will focus on developing a lightweight, learning-based matcher suitable for embedded devices under challenging scenarios.


\section*{Data Availability}

All the data used in this paper come from open-source datasets that are freely available for anyone to download. The demo and weights of our work are available at \url{https://github.com/HITCSC/EdgePoint2}.

\bibliographystyle{spbasic}      
\bibliography{ref}   

\begin{thebibliography}{72}
\providecommand{\natexlab}[1]{#1}
\providecommand{\url}[1]{{#1}}
\providecommand{\urlprefix}{URL }
\expandafter\ifx\csname urlstyle\endcsname\relax
  \providecommand{\doi}[1]{DOI~\discretionary{}{}{}#1}\else
  \providecommand{\doi}{DOI~\discretionary{}{}{}\begingroup \urlstyle{rm}\Url}\fi
\providecommand{\eprint}[2][]{\url{#2}}

\bibitem[{Arandjelovic et~al.(2016)Arandjelovic, Gronat, Torii, Pajdla, and Sivic}]{NetVLAD}
Arandjelovic R, Gronat P, Torii A, Pajdla T, Sivic J (2016) Netvlad: Cnn architecture for weakly supervised place recognition. In: Proceedings of the IEEE conference on computer vision and pattern recognition, pp 5297--5307

\bibitem[{Balntas et~al.(2017)Balntas, Lenc, Vedaldi, and Mikolajczyk}]{HPatches}
Balntas V, Lenc K, Vedaldi A, Mikolajczyk K (2017) Hpatches: A benchmark and evaluation of handcrafted and learned local descriptors. In: Proceedings of the IEEE conference on computer vision and pattern recognition, pp 5173--5182

\bibitem[{Barath et~al.(2020)Barath, Noskova, Ivashechkin, and Matas}]{MAGSAC}
Barath D, Noskova J, Ivashechkin M, Matas J (2020) Magsac++, a fast, reliable and accurate robust estimator. In: Proceedings of the IEEE/CVF conference on computer vision and pattern recognition, pp 1304--1312

\bibitem[{Bian et~al.(2019)Bian, Wu, Zhao, Liu, Zhang, Cheng, and Reid}]{FM-Bench}
Bian JW, Wu YH, Zhao J, Liu Y, Zhang L, Cheng MM, Reid I (2019) An evaluation of feature matchers for fundamental matrix estimation. arXiv preprint arXiv:190809474

\bibitem[{Calonder et~al.(2010)Calonder, Lepetit, Strecha, and Fua}]{BRIEF}
Calonder M, Lepetit V, Strecha C, Fua P (2010) Brief: Binary robust independent elementary features. In: Computer Vision--ECCV 2010: 11th European Conference on Computer Vision, Heraklion, Crete, Greece, September 5-11, 2010, Proceedings, Part IV 11, Springer, pp 778--792

\bibitem[{Campos et~al.(2021)Campos, Elvira, Rodr{\'\i}guez, Montiel, and Tard{\'o}s}]{ORBSLAM3}
Campos C, Elvira R, Rodr{\'\i}guez JJG, Montiel JM, Tard{\'o}s JD (2021) Orb-slam3: An accurate open-source library for visual, visual--inertial, and multimap slam. IEEE Transactions on Robotics 37(6):1874--1890

\bibitem[{Chen et~al.(2023)Chen, Kao, He, Zhuo, Wen, Lee, and Chan}]{FasterNet}
Chen J, Kao Sh, He H, Zhuo W, Wen S, Lee CH, Chan SHG (2023) Run, don't walk: Chasing higher flops for faster neural networks. In: Proceedings of the IEEE/CVF Conference on Computer Vision and Pattern Recognition, pp 12021--12031

\bibitem[{Christiansen et~al.(2019)Christiansen, Kragh, Brodskiy, and Karstoft}]{UnsuperPoint}
Christiansen PH, Kragh MF, Brodskiy Y, Karstoft H (2019) Unsuperpoint: End-to-end unsupervised interest point detector and descriptor. arXiv preprint arXiv:190704011

\bibitem[{Dai et~al.(2017)Dai, Chang, Savva, Halber, Funkhouser, and Nie{\ss}ner}]{ScanNet}
Dai A, Chang AX, Savva M, Halber M, Funkhouser T, Nie{\ss}ner M (2017) Scannet: Richly-annotated 3d reconstructions of indoor scenes. In: Proceedings of the IEEE conference on computer vision and pattern recognition, pp 5828--5839

\bibitem[{DeTone et~al.(2018)DeTone, Malisiewicz, and Rabinovich}]{SuperPoint}
DeTone D, Malisiewicz T, Rabinovich A (2018) Superpoint: Self-supervised interest point detection and description. In: Proceedings of the IEEE conference on computer vision and pattern recognition workshops, pp 224--236

\bibitem[{Duan et~al.(2017)Duan, Lu, Wang, Feng, and Zhou}]{DCBD-MQ}
Duan Y, Lu J, Wang Z, Feng J, Zhou J (2017) Learning deep binary descriptor with multi-quantization. In: Proceedings of the IEEE conference on computer vision and pattern recognition, pp 1183--1192

\bibitem[{Dusmanu et~al.(2019)Dusmanu, Rocco, Pajdla, Pollefeys, Sivic, Torii, and Sattler}]{D2-Net}
Dusmanu M, Rocco I, Pajdla T, Pollefeys M, Sivic J, Torii A, Sattler T (2019) D2-net: A trainable cnn for joint detection and description of local features. arXiv preprint arXiv:190503561

\bibitem[{Edstedt et~al.(2024{\natexlab{a}})Edstedt, Bökman, Wadenbäck, and Felsberg}]{DeDoDe}
Edstedt J, Bökman G, Wadenbäck M, Felsberg M (2024{\natexlab{a}}) {DeDoDe: Detect, Don't Describe --- Describe, Don't Detect for Local Feature Matching}. In: 2024 International Conference on 3D Vision (3DV), IEEE

\bibitem[{Edstedt et~al.(2024{\natexlab{b}})Edstedt, Bökman, and Zhao}]{DeDoDev2}
Edstedt J, Bökman G, Zhao Z (2024{\natexlab{b}}) {DeDoDe v2: Analyzing and Improving the DeDoDe Keypoint Detector }. In: IEEE/CVF Computer Society Conference on Computer Vision and Pattern Recognition Workshops (CVPRW)

\bibitem[{Fan et~al.(2022)Fan, Zhou, Feng, Pu, Yang, Kong, Wu, and Liu}]{SeLF}
Fan B, Zhou J, Feng W, Pu H, Yang Y, Kong Q, Wu F, Liu H (2022) Learning semantic-aware local features for long term visual localization. IEEE Transactions on Image Processing 31:4842--4855

\bibitem[{Fujimoto and Matsunaga(2023)}]{SPSG-Odometry}
Fujimoto S, Matsunaga N (2023) Deep feature-based rgb-d odometry using superpoint and superglue. Procedia Computer Science 227:1127--1134

\bibitem[{Geiger et~al.(2012)Geiger, Lenz, and Urtasun}]{KITTI}
Geiger A, Lenz P, Urtasun R (2012) Are we ready for autonomous driving? the kitti vision benchmark suite. In: 2012 IEEE conference on computer vision and pattern recognition, IEEE, pp 3354--3361

\bibitem[{Gleize et~al.(2023)Gleize, Wang, and Feiszli}]{SiLK}
Gleize P, Wang W, Feiszli M (2023) Silk: Simple learned keypoints. In: Proceedings of the IEEE/CVF international conference on computer vision, pp 22499--22508

\bibitem[{He et~al.(2016)He, Zhang, Ren, and Sun}]{ResNet}
He K, Zhang X, Ren S, Sun J (2016) Deep residual learning for image recognition. In: Proceedings of the IEEE conference on computer vision and pattern recognition, pp 770--778

\bibitem[{Howard et~al.(2019)Howard, Sandler, Chu, Chen, Chen, Tan, Wang, Zhu, Pang, Vasudevan et~al.}]{MobileNetv3}
Howard A, Sandler M, Chu G, Chen LC, Chen B, Tan M, Wang W, Zhu Y, Pang R, Vasudevan V, et~al. (2019) Searching for mobilenetv3. In: Proceedings of the IEEE/CVF international conference on computer vision, pp 1314--1324

\bibitem[{Howard et~al.(2022)Howard, Trulls, etru1927, Yi, old ufo, Dane, and Jin}]{IMC2022}
Howard A, Trulls E, etru1927, Yi KM, old ufo, Dane S, Jin Y (2022) Image matching challenge 2022. \urlprefix\url{https://kaggle.com/competitions/image-matching-challenge-2022}

\bibitem[{Kanakis et~al.(2023)Kanakis, Maurer, Spallanzani, Chhatkuli, and Van~Gool}]{ZippyPoint}
Kanakis M, Maurer S, Spallanzani M, Chhatkuli A, Van~Gool L (2023) Zippypoint: Fast interest point detection, description, and matching through mixed precision discretization. In: Proceedings of the IEEE/CVF Conference on Computer Vision and Pattern Recognition, pp 6113--6122

\bibitem[{Knapitsch et~al.(2017)Knapitsch, Park, Zhou, and Koltun}]{TT}
Knapitsch A, Park J, Zhou QY, Koltun V (2017) Tanks and temples: Benchmarking large-scale scene reconstruction. ACM Transactions on Graphics (ToG) 36(4):1--13

\bibitem[{Larsson and contributors(2020)}]{LO-RANSAC}
Larsson V, contributors (2020) {PoseLib - Minimal Solvers for Camera Pose Estimation}. \urlprefix\url{https://github.com/vlarsson/PoseLib}

\bibitem[{Li and Snavely(2018)}]{MegaDepth}
Li Z, Snavely N (2018) Megadepth: Learning single-view depth prediction from internet photos. In: Proceedings of the IEEE conference on computer vision and pattern recognition, pp 2041--2050

\bibitem[{Lin et~al.(2016)Lin, Lu, Chen, and Zhou}]{DeepBit}
Lin K, Lu J, Chen CS, Zhou J (2016) Learning compact binary descriptors with unsupervised deep neural networks. In: Proceedings of the IEEE conference on computer vision and pattern recognition, pp 1183--1192

\bibitem[{Lin et~al.(2014)Lin, Maire, Belongie, Hays, Perona, Ramanan, Doll{\'a}r, and Zitnick}]{COCO}
Lin TY, Maire M, Belongie S, Hays J, Perona P, Ramanan D, Doll{\'a}r P, Zitnick CL (2014) Microsoft coco: Common objects in context. In: Computer Vision--ECCV 2014: 13th European Conference, Zurich, Switzerland, September 6-12, 2014, Proceedings, Part V 13, Springer, pp 740--755

\bibitem[{Liu et~al.(2024)Liu, Feng, Xu, Ning, Xu, and Shen}]{OmniNXT}
Liu P, Feng C, Xu Y, Ning Y, Xu H, Shen S (2024) Omninxt: A fully open-source and compact aerial robot with omnidirectional visual perception. arXiv preprint arXiv:240320085

\bibitem[{Liu and Dong(2024)}]{DescriptorDistillation}
Liu Y, Dong Q (2024) Descriptor distillation: A teacher-student-regularized framework for learning local descriptors. International Journal of Computer Vision pp 1--19

\bibitem[{Loshchilov(2017)}]{AdamW}
Loshchilov I (2017) Decoupled weight decay regularization. arXiv preprint arXiv:171105101

\bibitem[{Lowe(2004)}]{SIFT}
Lowe DG (2004) Distinctive image features from scale-invariant keypoints. International journal of computer vision 60:91--110

\bibitem[{Luo et~al.(2020)Luo, Zhou, Bai, Chen, Zhang, Yao, Li, Fang, and Quan}]{ASLFeat}
Luo Z, Zhou L, Bai X, Chen H, Zhang J, Yao Y, Li S, Fang T, Quan L (2020) Aslfeat: Learning local features of accurate shape and localization. In: Proceedings of the IEEE/CVF conference on computer vision and pattern recognition, pp 6589--6598

\bibitem[{Ono et~al.(2018)Ono, Trulls, Fua, and Yi}]{LF-Net}
Ono Y, Trulls E, Fua P, Yi KM (2018) Lf-net: Learning local features from images. Advances in neural information processing systems 31

\bibitem[{Pan et~al.(2024)Pan, Baráth, Pollefeys, and Sch\"{o}nberger}]{glomap}
Pan L, Baráth D, Pollefeys M, Sch\"{o}nberger JL (2024) Global structure-from-motion revisited. In: European Conference on Computer Vision (ECCV)

\bibitem[{Potje et~al.(2024)Potje, Cadar, Araujo, Martins, and Nascimento}]{XFeat}
Potje G, Cadar F, Araujo A, Martins R, Nascimento ER (2024) Xfeat: Accelerated features for lightweight image matching. In: Proceedings of the IEEE/CVF Conference on Computer Vision and Pattern Recognition, pp 2682--2691

\bibitem[{Qin et~al.(2018)Qin, Li, and Shen}]{VINS-Mono}
Qin T, Li P, Shen S (2018) Vins-mono: A robust and versatile monocular visual-inertial state estimator. IEEE Transactions on Robotics 34(4):1004--1020

\bibitem[{Revaud et~al.(2019)Revaud, Weinzaepfel, De~Souza, Pion, Csurka, Cabon, and Humenberger}]{R2D2}
Revaud J, Weinzaepfel P, De~Souza C, Pion N, Csurka G, Cabon Y, Humenberger M (2019) R2d2: repeatable and reliable detector and descriptor. arXiv preprint arXiv:190606195

\bibitem[{Rosten and Drummond(2006)}]{FAST}
Rosten E, Drummond T (2006) Machine learning for high-speed corner detection. In: Computer Vision--ECCV 2006: 9th European Conference on Computer Vision, Graz, Austria, May 7-13, 2006. Proceedings, Part I 9, Springer, pp 430--443

\bibitem[{Rublee et~al.(2011)Rublee, Rabaud, Konolige, and Bradski}]{ORB}
Rublee E, Rabaud V, Konolige K, Bradski G (2011) Orb: An efficient alternative to sift or surf. In: 2011 International conference on computer vision, Ieee, pp 2564--2571

\bibitem[{Sandler et~al.(2018)Sandler, Howard, Zhu, Zhmoginov, and Chen}]{MobileNetv2}
Sandler M, Howard A, Zhu M, Zhmoginov A, Chen LC (2018) Mobilenetv2: Inverted residuals and linear bottlenecks. In: Proceedings of the IEEE conference on computer vision and pattern recognition, pp 4510--4520

\bibitem[{Sarlin et~al.(2019{\natexlab{a}})Sarlin, Cadena, Siegwart, and Dymczyk}]{HF-Net}
Sarlin PE, Cadena C, Siegwart R, Dymczyk M (2019{\natexlab{a}}) From coarse to fine: Robust hierarchical localization at large scale. In: Proceedings of the IEEE/CVF Conference on Computer Vision and Pattern Recognition, pp 12716--12725

\bibitem[{Sarlin et~al.(2019{\natexlab{b}})Sarlin, Cadena, Siegwart, and Dymczyk}]{HLoc}
Sarlin PE, Cadena C, Siegwart R, Dymczyk M (2019{\natexlab{b}}) From coarse to fine: Robust hierarchical localization at large scale. In: CVPR

\bibitem[{Sattler et~al.(2018)Sattler, Maddern, Toft, Torii, Hammarstrand, Stenborg, Safari, Okutomi, Pollefeys, Sivic et~al.}]{HLocDataset}
Sattler T, Maddern W, Toft C, Torii A, Hammarstrand L, Stenborg E, Safari D, Okutomi M, Pollefeys M, Sivic J, et~al. (2018) Benchmarking 6dof outdoor visual localization in changing conditions. In: Proceedings of the IEEE conference on computer vision and pattern recognition, pp 8601--8610

\bibitem[{Schonberger and Frahm(2016)}]{colmap1}
Schonberger JL, Frahm JM (2016) Structure-from-motion revisited. In: Proceedings of the IEEE conference on computer vision and pattern recognition, pp 4104--4113

\bibitem[{Sch{\"o}nemann(1966)}]{OPP}
Sch{\"o}nemann PH (1966) A generalized solution of the orthogonal procrustes problem. Psychometrika 31(1):1--10

\bibitem[{Simonyan and Zisserman(2014)}]{VGG}
Simonyan K, Zisserman A (2014) Very deep convolutional networks for large-scale image recognition. arXiv preprint arXiv:14091556

\bibitem[{Song et~al.(2015)Song, Chandraker, and Guest}]{song2015high}
Song S, Chandraker M, Guest CC (2015) High accuracy monocular sfm and scale correction for autonomous driving. IEEE transactions on pattern analysis and machine intelligence 38(4):730--743

\bibitem[{Sturm et~al.(2012)Sturm, Engelhard, Endres, Burgard, and Cremers}]{TUM}
Sturm J, Engelhard N, Endres F, Burgard W, Cremers D (2012) A benchmark for the evaluation of rgb-d slam systems. In: 2012 IEEE/RSJ international conference on intelligent robots and systems, IEEE, pp 573--580

\bibitem[{Sun et~al.(2021)Sun, Shen, Wang, Bao, and Zhou}]{LoFTR}
Sun J, Shen Z, Wang Y, Bao H, Zhou X (2021) Loftr: Detector-free local feature matching with transformers. In: Proceedings of the IEEE/CVF conference on computer vision and pattern recognition, pp 8922--8931

\bibitem[{Taira et~al.(2018)Taira, Okutomi, Sattler, Cimpoi, Pollefeys, Sivic, Pajdla, and Torii}]{InLoc}
Taira H, Okutomi M, Sattler T, Cimpoi M, Pollefeys M, Sivic J, Pajdla T, Torii A (2018) Inloc: Indoor visual localization with dense matching and view synthesis. In: Proceedings of the IEEE Conference on Computer Vision and Pattern Recognition, pp 7199--7209

\bibitem[{Tan and \vspace{0mm} Le(2021)}]{EfficientNetv2}
Tan M, \vspace{0mm} Le Q (2021) Efficientnetv2: Smaller models and faster training. In: International conference on machine learning, PMLR, pp 10096--10106

\bibitem[{Tang et~al.(2019)Tang, Kim, Guizilini, Pillai, and Ambrus}]{KP2D}
Tang J, Kim H, Guizilini V, Pillai S, Ambrus R (2019) Neural outlier rejection for self-supervised keypoint learning. arXiv preprint arXiv:191210615

\bibitem[{Tian et~al.(2019)Tian, Yu, Fan, Wu, Heijnen, and Balntas}]{SOSNet}
Tian Y, Yu X, Fan B, Wu F, Heijnen H, Balntas V (2019) Sosnet: Second order similarity regularization for local descriptor learning. In: Proceedings of the IEEE/CVF Conference on Computer Vision and Pattern Recognition, pp 11016--11025

\bibitem[{Tian et~al.(2022)Tian, Chang, Arias, Nieto-Granda, How, and Carlone}]{KimeraMulti2022}
Tian Y, Chang Y, Arias FH, Nieto-Granda C, How JP, Carlone L (2022) Kimera-multi: Robust, distributed, dense metric-semantic slam for multi-robot systems. IEEE Transactions on Robotics 38(4)

\bibitem[{Tung and Mori(2019)}]{Similarity-Preserving}
Tung F, Mori G (2019) Similarity-preserving knowledge distillation. In: Proceedings of the IEEE/CVF international conference on computer vision, pp 1365--1374

\bibitem[{Tyszkiewicz et~al.(2020)Tyszkiewicz, Fua, and Trulls}]{DISK}
Tyszkiewicz M, Fua P, Trulls E (2020) Disk: Learning local features with policy gradient. Advances in Neural Information Processing Systems 33:14254--14265

\bibitem[{Wang et~al.(2023)Wang, Xu, Lu, Xu, Meng, Zhang, Fan, and Zhang}]{AWDesc}
Wang C, Xu R, Lu K, Xu S, Meng W, Zhang Y, Fan B, Zhang X (2023) Attention weighted local descriptors. IEEE Transactions on Pattern Analysis and Machine Intelligence 45(9):10632--10649

\bibitem[{Wang et~al.(2024{\natexlab{a}})Wang, Jia, Wang, and Zuo}]{wang2024relation}
Wang H, Jia T, Wang Q, Zuo W (2024{\natexlab{a}}) Relation knowledge distillation by auxiliary learning for object detection. IEEE Transactions on Image Processing

\bibitem[{Wang et~al.(2024{\natexlab{b}})Wang, Wu, Yang, Zuo, and Hu}]{wang2024layer}
Wang Q, Wu Y, Yang L, Zuo W, Hu Q (2024{\natexlab{b}}) Layer-specific knowledge distillation for class incremental semantic segmentation. IEEE Transactions on Image Processing

\bibitem[{Wang et~al.(2024{\natexlab{c}})Wang, He, Peng, Tan, and Zhou}]{EfficientLoFTR}
Wang Y, He X, Peng S, Tan D, Zhou X (2024{\natexlab{c}}) Efficient loftr: Semi-dense local feature matching with sparse-like speed. In: Proceedings of the IEEE/CVF Conference on Computer Vision and Pattern Recognition, pp 21666--21675

\bibitem[{Wang et~al.(2024{\natexlab{d}})Wang, Ng, Sa, Parra, Rodriguez-Opazo, Lin, and Li}]{MAVIS}
Wang Y, Ng Y, Sa I, Parra A, Rodriguez-Opazo C, Lin T, Li H (2024{\natexlab{d}}) Mavis: Multi-camera augmented visual-inertial slam using se 2 (3) based exact imu pre-integration. In: 2024 IEEE International Conference on Robotics and Automation (ICRA), IEEE, pp 1694--1700

\bibitem[{Wang et~al.(2022)Wang, Xiao, Duan, Zhou, and Lu}]{D-GraphBit}
Wang Z, Xiao H, Duan Y, Zhou J, Lu J (2022) Learning deep binary descriptors via bitwise interaction mining. IEEE Transactions on Pattern Analysis and Machine Intelligence 45(2):1919--1933

\bibitem[{Wilson and Snavely(2014)}]{CPC}
Wilson K, Snavely N (2014) Robust global translations with 1dsfm. In: Computer Vision--ECCV 2014: 13th European Conference, Zurich, Switzerland, September 6-12, 2014, Proceedings, Part III 13, Springer, pp 61--75

\bibitem[{Xu et~al.(2022{\natexlab{a}})Xu, Liu, Chen, and Shen}]{D2SLAM}
Xu H, Liu P, Chen X, Shen S (2022{\natexlab{a}}) {$D^{2}$ SLAM}: Decentralized and distributed collaborative visual-inertial slam system for aerial swarm. arXiv preprint arXiv:221101538

\bibitem[{Xu et~al.(2022{\natexlab{b}})Xu, Zhang, Zhou, Wang, Yao, Meng, and Shen}]{OmniSwarm}
Xu H, Zhang Y, Zhou B, Wang L, Yao X, Meng G, Shen S (2022{\natexlab{b}}) Omni-swarm: A decentralized omnidirectional visual--inertial--uwb state estimation system for aerial swarms. IEEE Transactions on Robotics 38(6):3374--3394

\bibitem[{Yang et~al.(2022{\natexlab{a}})Yang, Li, Jiang, Gong, Yuan, Zhao, and Yuan}]{yang2022focal}
Yang Z, Li Z, Jiang X, Gong Y, Yuan Z, Zhao D, Yuan C (2022{\natexlab{a}}) Focal and global knowledge distillation for detectors. In: Proceedings of the IEEE/CVF Conference on Computer Vision and Pattern Recognition, pp 4643--4652

\bibitem[{Yang et~al.(2022{\natexlab{b}})Yang, Li, Shao, Shi, Yuan, and Yuan}]{yang2022masked}
Yang Z, Li Z, Shao M, Shi D, Yuan Z, Yuan C (2022{\natexlab{b}}) Masked generative distillation. In: European Conference on Computer Vision, Springer, pp 53--69

\bibitem[{Yao et~al.(2024)Yao, Hao, Xie, and He}]{EdgePoint}
Yao H, Hao N, Xie C, He F (2024) Edgepoint: Efficient point detection and compact description via distillation. In: 2024 IEEE International Conference on Robotics and Automation (ICRA), IEEE, pp 766--772

\bibitem[{Yin et~al.(2021)Yin, Li, Li, Yu, and Zou}]{M2DGR}
Yin J, Li A, Li T, Yu W, Zou D (2021) M2dgr: A multi-sensor and multi-scenario slam dataset for ground robots. IEEE Robotics and Automation Letters 7(2):2266--2273

\bibitem[{Yin et~al.(2023)Yin, Feng, Fan, Ju, and Zhang}]{SP-VSLAM}
Yin Z, Feng D, Fan C, Ju C, Zhang F (2023) Sp-vslam: Monocular visual-slam algorithm based on superpoint network. In: 2023 15th International Conference on Communication Software and Networks (ICCSN), IEEE, pp 456--459

\bibitem[{Zhao et~al.(2022)Zhao, Wu, Miao, Chen, Chen, and Li}]{ALIKE}
Zhao X, Wu X, Miao J, Chen W, Chen PC, Li Z (2022) Alike: Accurate and lightweight keypoint detection and descriptor extraction. IEEE Transactions on Multimedia 25:3101--3112

\bibitem[{Zhao et~al.(2023)Zhao, Wu, Chen, Chen, Xu, and Li}]{ALIKED}
Zhao X, Wu X, Chen W, Chen PC, Xu Q, Li Z (2023) Aliked: A lighter keypoint and descriptor extraction network via deformable transformation. IEEE Transactions on Instrumentation and Measurement 72:1--16

\end{thebibliography}

\end{document}